\documentclass[runningheads]{llncs}

\usepackage{eccv}

\usepackage{eccvabbrv}

\usepackage{graphicx}
\usepackage{booktabs}
\usepackage{algorithm}
\usepackage{algorithmic}
\usepackage{wrapfig}

\usepackage[accsupp]{axessibility}  

\usepackage{hyperref}

\makeatletter
\newcommand{\printfnsymbol}[1]{%
  \textsuperscript{\@fnsymbol{#1}}%
}
\makeatother

\usepackage{orcidlink}
\usepackage{graphicx}
\usepackage{amsmath}
\usepackage{amssymb}
\usepackage{booktabs}
\usepackage{xcolor}
\usepackage{color, colortbl}
\usepackage{pifont}
\usepackage{nicematrix}

\usepackage[capitalize]{cleveref}

\definecolor{bluex}{rgb}{0.27, 0.42, 0.81}
\definecolor{purplex}{HTML}{9564bf}
\definecolor{red3}{HTML}{C52A20}
\definecolor{red2}{HTML}{B36A6F}
\definecolor{red1}{HTML}{FFb5b5}
\definecolor{purple}{HTML}{B36A6F}
\definecolor{darkyellow}{HTML}{D5BA82}
\definecolor{blue1}{HTML}{508AB2}
\definecolor{blue2}{HTML}{C4E4E3}
\definecolor{green1}{HTML}{A1D0C7}
\definecolor{green2}{HTML}{BFF6BA}
\definecolor{green3}{HTML}{028100}
\definecolor{teal}{HTML}{508AB2}
\definecolor{Gray}{gray}{0.94}

\usepackage[listings]{tcolorbox}
\tcbuselibrary{listings,theorems}
\newtcolorbox{mybox}{colback=white!5!white,colframe=black!75!black, left=.05in, right=.05in}

\newtcbtheorem{emptybox}{}%
{colback=green2!5,colframe=blue1,fonttitle=\bfseries,theorem style=plain, left=.0in, right=.0in,bottom=.02in, top=.02in,terminator sign none}{emptybox}
\newtcbtheorem{bth}{Theoreme}
{colback=red!20,colframe=red,theorem style=plain,fonttitle=\bfseries,coltitle=black,terminator sign none}{th}

\newtcbtheorem{exmp}{Example}%
{colback=green2!5,colframe=blue1,fonttitle=\bfseries, left=.02in, right=.02in,bottom=.02in, top=.02in}{ex}
\newtcbtheorem[]{prompt}{Backward Prompting}%
{colback=green!5,colframe=green!35!black,fonttitle=\bfseries}{th}
\newtcbtheorem[number within=section]{thm}{Theorem}%
{colback=green!5,colframe=green!35!black,fonttitle=\bfseries}{th}
\newtcbtheorem[number within=section]{corr}{Corollary}%
{colback=green!5,colframe=green!35!black,fonttitle=\bfseries}{th}
\newtcbtheorem{method}{Method}%
{colback=green!5,colframe=green!35!black,fonttitle=\bfseries}{th}

\newtcbtheorem{ebox}{}
{colback=green2!5,colframe=blue1, left=.02in, right=.02in,bottom=.02in,title empty, top=.02in, theorem style=plain apart,frame empty}{ebox}

\hypersetup{linkcolor=[rgb]{0.7,0.1,0.1}}
\hypersetup{citecolor=[HTML]{1663A9}}

\crefname{section}{Sec.}{Secs.}
\Crefname{section}{Section}{Sections}
\Crefname{table}{Table}{Tables}
\crefname{table}{Tab.}{Tabs.}

\definecolor{Gray}{gray}{0.94}
\definecolor{Grayb}{gray}{0.88}

\newcommand{\cmark}{\ding{51}}%
\newcommand{\xmark}{\ding{55}}

\newcommand{\vx}{{\bf x}}
\newcommand{\bT}{\mathbb{T}}

\newcommand{\hD}{\mathcal{D}}
\newcommand{\vtheta}{{\boldsymbol \theta}}

\begin{document}

\title{VLLaVO: Mitigating Visual Gap through LLMs}


\author{Shuhao~Chen\inst{1}\thanks{Equal contribution.} \and
Yulong~Zhang\inst{2}\printfnsymbol{1} \and
Weisen~Jiang\inst{1,3} \and 
Jiangang~Lu\inst{2} \and
Yu~Zhang\inst{1}\thanks{Corresponding author.}
}

\authorrunning{Chen et al.}

\institute{Southern University of Science and Technology \and
Zhejiang University \and 
Hong Kong University of Science and Technology \\
Project Page: \url{https://ll-a-vo.github.io/}}

\maketitle

\begin{abstract}
Recent advances achieved by deep learning models rely on the independent and identically distributed assumption, hindering their applications in real-world scenarios with domain shifts. 
To tackle this issue, cross-domain learning aims at extracting domain-invariant knowledge to reduce the domain shift between training and testing data.
However, in visual cross-domain learning, traditional methods concentrate solely on the image modality, disregarding the potential benefits of incorporating the text modality.
In this work, we propose VLLaVO, combining Vision language models and Large Language models as Visual cross-dOmain learners.
VLLaVO uses vision-language models to convert images into detailed textual descriptions. 
A large language model is then finetuned on textual descriptions of the source/target domain generated by a designed instruction template.
Extensive experimental results 
under domain generalization and unsupervised domain adaptation settings demonstrate the effectiveness of the proposed method.
  \keywords{Domain Generalization \and Unsupervised Domain Adaptation \and Large Language Models}
\end{abstract}

\section{Introduction}
\label{sec:intro}

Deep models have achieved promising performance in various applications of computer vision
when testing images are sampled from the same distribution of training data~\cite{he2016deep, dosovitskiy2020image, rahman2023learning, lin2023dynamicdet}.
However, in real-world applications, domain shifts \cite{yang2020transfer} remain a significant challenge due to the distribution discrepancy between training and testing domains.
For instance, an autonomous driving car should be able to handle adverse weather which does not appear during the training stage~\cite{dai2018dark, michaelis2019benchmarking, sun2022shift},
posing a substantial hurdle for model generalization.
To alleviate the domain shift, cross-domain learning~\cite{xia2020structure, zhang2022adaptive, xu2022cdtrans} aims at extracting domain-invariant knowledge between source and target domains.

\begin{figure}[!t]
\centering
\!\!\!\!\!\!
\subcaptionbox{
CLIP embeddings.
\label{fig:tsne_image_demo}
}
{\includegraphics[width=0.35\textwidth]{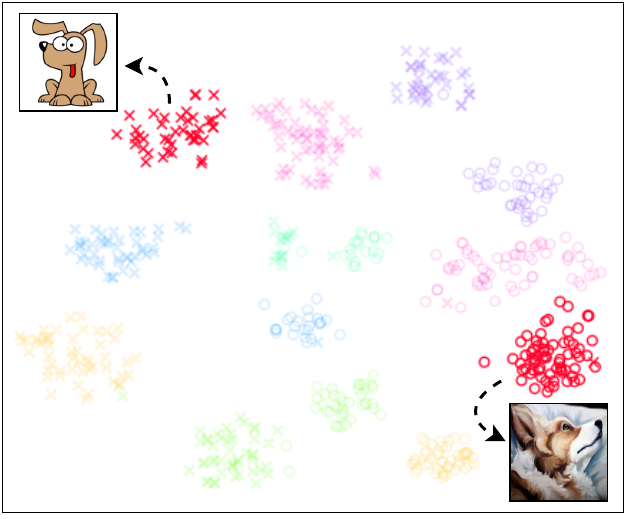}}
\quad\;
\subcaptionbox{
LLM embeddings.
\label{fig:tsne_text_demo}
} {\includegraphics[width=0.35\textwidth]{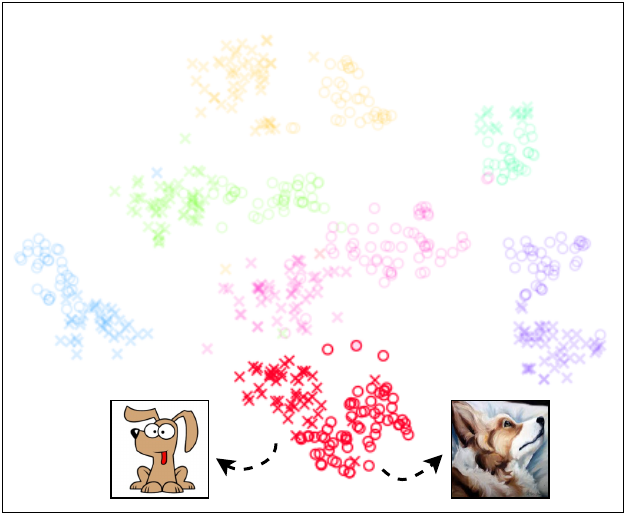}}
\!\!\!\!\!\!
\caption{t-SNE visualizations of samples from the Art Painting domain (marked by ``\textsf{o}'') and the Cartoon domain (marked by ``\textsf{x}'') on the \textit{PACS} dataset. Samples of the same category are depicted in the same color. Implementation details are provided in Section~\ref{sec:domain_invariant}.}
\label{fig:tsne_demo}
\vspace{-0.5cm} 
\end{figure}

Typical cross-domain learning scenarios include domain generalization (DG)~\cite{cha2021swad, wang2023sharpness, shu2023clipood} and unsupervised domain adaptation (UDA)~\cite{zhang2023free, xu2022cdtrans, lai2023padclip, zhang2023domainx}.
Specifically, DG trains a model from multiple source domains and evaluates it on an unseen target domain,
while UDA assumes some unlabeled data
from the target domain is available for training.
Various DG \cite{cha2022domain, li2018deep, zhou2021domain} and UDA \cite{xu2022cdtrans, zhang2023free, long2015learning} methods have been proposed to 
alleviate the domain shift by learning domain-invariant features in the image modality.
For example, under the DG setting, MIRO~\cite{cha2022domain} proposes to learn features that are similar to the oracle representation. 
For the UDA setting, 
BCAT~\cite{wang2023unsupervised} leverages a bidirectional cross-attention mechanism to extract implicit source and target mixup feature representations. 
However, the above methods only utilize the image modality for cross-domain learning.

Instead of just using the image modality, recent efforts on large-scale vision-language models (VLMs)~\cite{radford2021learning, zhou2022learning, gao2023clip} have shown significant improvements in image classification by learning from massive paired image-text samples. For example, CLIP~\cite{radford2021learning}, which consists of an image encoder and a text encoder, is trained on 400 million image-text pairs by contrastive learning~\cite{chen2020simple}. However, as shown in Figure~\ref{fig:tsne_image_demo}, learning domain-invariant features based on CLIP is challenging due to the large domain shift, leading to sub-optimal cross-domain performance. 
On the other hand, large language models (LLMs) \cite{touvron2023llama, brown2020language} have impressive zero-shot capability for text-based tasks. Nevertheless, it is essential to acknowledge that the success of LLMs does not readily extend to pure vision and vision-language tasks, primarily due to the inherent disparities in modalities and task structures. To the best of our knowledge, there is no work that integrates LLMs with visual cross-domain learning.

In this paper,
we propose VLLaVO, combining \textbf{V}ision language models and \textbf{L}arge \textbf{La}nguage models as \textbf{V}isual cross-d\textbf{O}main learners, to integrate the powerful LLMs into visual cross-domain learning.
Firstly, we use various VLMs (including CLIP \cite{radford2021learning} and BLIP \cite{li2022blip}) to convert images into detailed textual descriptions (\textit{i.e.}, tags, attributes, and captions). 
As shown in Example \ref{ex:exmp1} placed in the experimental section, these descriptions contain crucial image content information. 
As a byproduct, based on these descriptions, we release a series of datasets in text modality for text classification tasks and the details are given in Appendix \ref{sec:text_dataset}.
Next, we design an instruction template for 
querying the LLM to obtain images' categories based on their descriptions.
However, the domain shift of generated textual descriptions still exists across domains.
A straightforward method \textit{rank classification} \cite{brown2020language} 
employs the output probability of LLM for each category to perform classification.
Empirical results indicate that this simple method performs poorly in challenging cross-domain tasks due to distractions from irrelevant contexts~\cite{shi2023large, yoran2023making, creswell2023selectioninference} and the issue of hallucinations~\cite{wan2023reformulating, zhang2023siren, sun2023contrastive}.
Especially in cross-domain learning scenarios, there are many domain-specific but class-irrelevant contexts \cite{yang2020transfer}.
To deal with these challenges, we finetune the LLM using training data comprised of question-answer pairs, where the questions are created by the proposed template and the answers correspond to the class tokens of corresponding images. 
To better adapt VLLaVO in the UDA scenario, we utilize the available target domain data with its pseudo-labels to further finetune the LLM. 
Experiments conducted on benchmark datasets under the DG and UDA settings 
have demonstrated that the proposed VLLaVO method 
achieves state-of-the-art performance.
Moreover, compared with existing VLM-based methods,
VLLaVO performs better, showing the benefits of integrating LLM into cross-domain learning.
    Additional support can be gained from the smaller domain shift of LLM than that of VLM (\textit{i.e.}, CLIP), as shown in Figure~\ref{fig:tsne_demo}.

Our contributions are summarized as follows.
\begin{itemize}
    \item We are the first to utilize LLMs for visual cross-domain learning. The proposed VLLaVO uses VLMs for bridging image and text modalities and designs a question instruction template for the LLM to predict the category of an image based on its description.
    \item The introduced VLLaVO alleviates domain shifts through the inherent generalization capability of LLM. Finetuning LLM on the designed question instruction template enhances its instruction-following ability, reducing the distractions of irrelevant contexts.
    \item Extensive experiments on diverse DG and UDA tasks demonstrate that VLLaVO consistently achieves state-of-the-art performance, surpassing existing VLM-based methods. 
    \item To facilitate further research, we release a series of datasets in the text modality, including 
    \textit{PACS}, \textit{OfficeHome}, and \textit{DomainNet}.
\end{itemize}

\section{Related Work}

\begin{figure*}[!t]
    \centering
    \includegraphics[width=\textwidth]{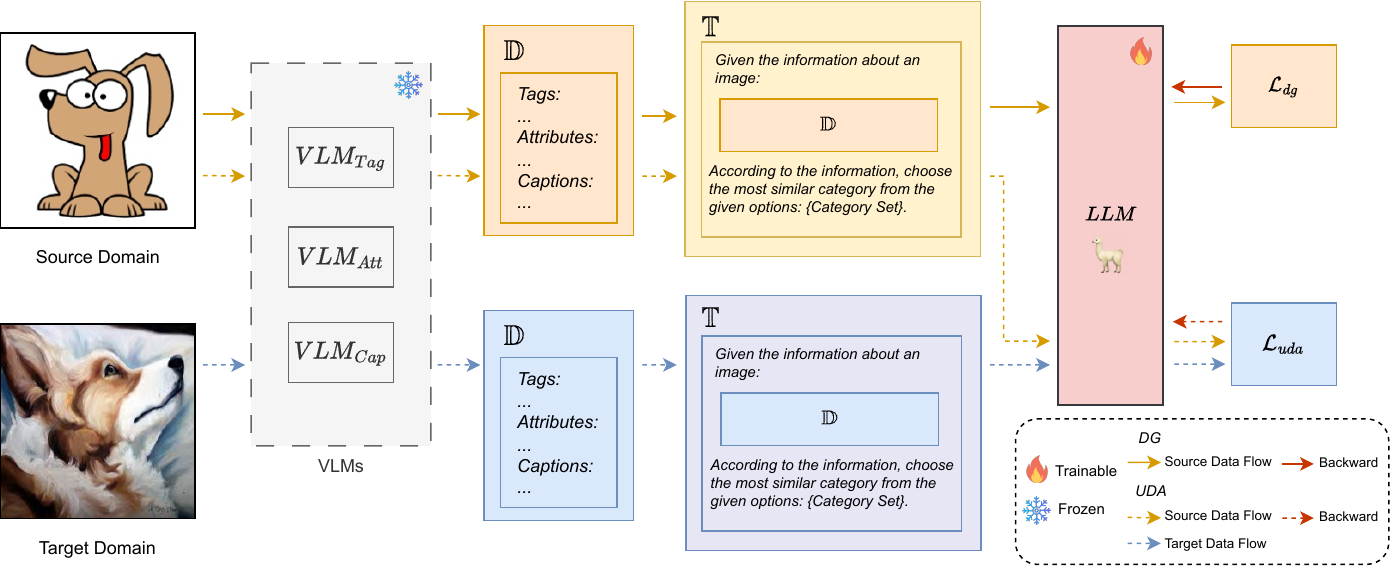} \vskip -.10in
    \caption{An illustration of the proposed VLLaVO framework for both UDA and DG. For DG, the source images are converted into the text modality through the VLMs. Then, the extracted descriptions are used to finetune an LLM. For UDA, we use the target domain samples with pseudo-labels and the source domain samples with ground truth labels to finetune the LLM.
    }    \label{fig:overall_method}
    \vspace{-0.4cm} 
\end{figure*}

The goal of cross-domain learning \cite{xia2020structure, zhang2022adaptive} is to learn a model from the source domain(s) with the possible help of the target domain to learn domain-invariant features and apply them to the target domain. 
Domain generalization and unsupervised domain adaptation are two typical scenarios in cross-domain learning.

\noindent\textbf{Domain Generalization (DG)}. DG \cite{cha2021swad, wang2023sharpness, cha2022domain, zhou2022domain} aims to improve the generalization capabilities of the models in new domains that are unseen during the training phase. DG is widely used in real-world applications \cite{zhou2022domain, dubey2021adaptive} and various methods have been proposed.
For example, SWAD \cite{cha2021swad} and SAGM \cite{wang2023sharpness} find flatter minima to improve the generalization capabilities.
Instead of learning domain-invariant features, MIRO \cite{cha2022domain} is proposed to learn features that are similar to the oracle representation.

\noindent \textbf{Unsupervised Domain Adaptation (UDA)}. Besides a source domain with labeled data, in the UDA scenario \cite{jin2020minimum, liu2021swin, zhang2023free}, unlabeled data in the target domain are available during the training phase.
CDTrans \cite{xu2022cdtrans} proposes a weight-sharing triple-branch transformer framework to enhance feature alignment and improve robustness to noisy pseudo-labels.
From a perspective of game theory, \mbox{PMTrans}~\cite{liu2021swin} builds up an intermediate domain with an effective ViT \cite{dosovitskiy2020image}-based solution to smoothly bridge the source and target domains.
Instead of explicitly aligning the source and target domains, MCC \cite{jin2020minimum} introduces a general class confusion loss as a regularizer.
ELS \cite{zhang2023free} adopts the domain adversarial training strategy, and mitigates the impact of label noise by encouraging the domain discriminator to output soft domain labels.

However, most existing DG and UDA methods focus solely on the image modality, leading to sub-optimal cross-domain effects as shown in our experiments. On the contrary, our VLLaVO further leverages the text modalities.

\noindent \textbf{VLMs and LLMs.}
VLMs have shown promising performance in learning visual representations \cite{jia2021scaling, menon2023visual, zhou2022conditional, alayrac2022flamingo}. A notable example is CLIP \cite{radford2021learning}, which is trained on 400 million image-text pairs and has been applied to various downstream tasks~\cite{gao2023clip, wang2023exploring, zhang2023domain}.
Recently, some works have applied CLIP to address domain shift problems \cite{ge2022domain, cho2023promptstyler}.
For example, DPL \cite{zhang2023domain} learns a lightweight prompt generator to generate a domain-specific prompt for efficiently adapting CLIP to an unseen domain.
CLIPood \cite{shu2023clipood} further leverages the semantic relationships between classes in the text modality and proposes a novel model optimization strategy to improve the generalization performance.
PADCLIP \cite{lai2023padclip} proposes a catastrophic forgetting metric to adjust the learning rate and alleviate the catastrophic forgetting issue of CLIP.
However, finetuning CLIP directly leads to a sacrifice in cross-domain generalization capability \cite{shu2023clipood, lai2023padclip}. 

On the other hand, LLMs show impressive generalization capability in various natural language processing tasks, \textit{e.g.}, topic classification \cite{ jiang2023effective, brown2020language}, mathematical reasoning \cite{wei2022chain, zhao2022jiuzhang, yu2023metamath, jiang2023backward}, and question answering \cite{brown2020language, wei2021finetuned, zhang2023multimodal}.
Moreover, instruction-tuning methods \cite{wei2021finetuned, chung2022scaling, ouyang2022training} further improve the zero-shot and few-shot performance of LLM.
Despite these advancements, LLMs cannot be directly applied to pure vision tasks and vision-language tasks due to the inherent disparities in modalities and task structure.
To address this issue, some works have utilized VLMs as a bridge between the image and text modalities.
For example, Menon et al.~\cite{menon2023visual} leverage an LLM (\textit{i.e.}, GPT-3 \cite{brown2020language}) to obtain descriptions for each category and employ CLIP to make predictions based on both the images and the extracted textual descriptions. 
LENS~\cite{berrios2023towards} designs a set of highly descriptive visual modules based on VLMs, enabling LLMs to tackle visual tasks.
Different from those works that assume that all the data are in the same domain, the proposed VLLaVO focuses on visual cross-domain learning.

\section{Methodology}

\subsection{Overview of The Entire Framework}

Suppose there is a set of $S$ source domains $\{\mathcal{D}_i\}_{i=1}^S$,
where $\mathcal{D}_i=\{(\vx_j^i,y_j^i)\}_{j=1}^{n_i}$ and $n_i$ denotes the number of samples in $\mathcal{D}_i$.
Let $\mathcal{D}=\cup_{i=1}^S \mathcal{D}_i$ denote the entire source domain training data.
In cross-domain learning, we aim to learn a model from source domains and then generalize the model to a target domain $\mathcal{T}$ whose data distribution differs from that of the source domains.
In DG, target domain data are unavailable during the training process, while in the training process of UDA, the unlabeled target domain data are available.
In UDA, $S$ equals to 1.

To begin with, {VLLaVO} generates textual descriptions of images using pre-trained VLMs. 
These obtained textual descriptions still exhibit domain shifts.
Thus, we finetune an LLM with the designed question instruction template that combines these descriptions and their category labels.
Finetuning enables the LLM to follow the question instruction template
and focus on the domain-invariant information related to the classification.
The overall framework of VLLaVO is shown in Figure \ref{fig:overall_method} and the entire algorithm is shown in Algorithm \ref{alg:train_pipeline} of Appendix \ref{sec:algorithm}. In the following sections, we will detail each step individually.

\subsection{Generating Textual Descriptions for Images}
\label{sec:extract}
    
For an image $\vx$, we use VLMs to convert it into textual descriptions. 
Previous works \cite{menon2023visual, yan2023learning} have demonstrated that textual descriptions can provide information for classification.
Following \cite{berrios2023towards}, we use pretrained VLMs to extract textual descriptions from three aspects: (i) Tags; (ii) Attributes; and (iii) Captions. 

To extract tags, a comprehensive tag vocabulary is constructed by integrating diverse sources, including multiple datasets such as \cite{russakovsky2015imagenet,li2022marc,cimpoi2014describing,parkhi2012cats,nilsback2008automated,bossard2014food,xiao2010sun,krause20133d,gupta2019lvis,lin2014microsoft,kuznetsova2020open,krishna2017visual}. 
Specifically, the tags are wrapped by a standardized template ``A photo of \{tag\}'' and then fed to the text encoder of the CLIP model to obtain the embeddings. 
For each image $\vx$, we compute the cosine similarity between its embedding extracted by the image encoder of the CLIP model and tags' embeddings.
The Top-$K$ similar tags are then used in the textual description.
    
However, tags are usually too brief to contain detailed descriptions of images.
To deal with this issue, we use GPT-3~\cite{brown2020language} to enrich the tags with attribute information.
Specifically, for each tag, we prompt GPT-3 with the input ``What are useful features for distinguishing a \{tag\} in a photo?'' to generate the crucial attributes of visual characteristics for each tag.
Similarly, all attributes are fed into the text encoder of the CLIP model to obtain the embeddings. 
For each image $\vx$, we first calculate the cosine similarity between the embedding of $\vx$, extracted by the image encoder of the CLIP model, and the embeddings of the attributes.
Then the Top-$M$ attributes with the highest similarity scores are used in the textual description.

As tags and attributes are concept-level descriptions of images, they lack specific content of images. To this end, we further enrich the textual information by extracting captions of images.
We use an image captioning model (\textit{e.g.}, BLIP \cite{li2022blip}) to generate Top-$N$ captions for each image by stochastic sampling \cite{fan2018hierarchical}.
    
Based on the extracted tags, attributes, and captions as described above, we can obtain the textual descriptions as
\begin{align*}
\mathbb{D}(\vx) = &\text{``Tags: \{Top-$K$ tags\}} \\
&\;\text{Attributes: \{Top-$M$ attributes\}} \\
&\;\text{Captions: \{Top-$N$ captions\}''}. 
\end{align*}
For illustration, Example~\ref{ex:exmp1} shows an example image and its corresponding extracted description.

\subsection{Question Instruction}
\label{sec:qa_prompt_pair}
    
After extracting textual descriptions from images,
we can use LLM for classification.
Specifically, we design a question instruction template for querying the LLM to its category.
The template $\bT(\vx)$ is shown in Example~\ref{ex:template},
where $\{\mathbb{D}(\vx)\}$ denotes the placeholder of the textual description of $\vx$ and \{\texttt{Category Set}\} is the placeholder of the candidate categories.
An example of the category set is shown in Appendix \ref{appendix:dataset}.
\begin{exmp}{Template $\bT(\vx)$.}{template}
\label{exmp_template1}
Give the information about an image:
\{$\mathbb{D}(\vx)$\}. \\
According to the information, choose the most similar category from the given options: 
\{\texttt{Category Set}\}. \\ 
\#\#\# Answer:
\end{exmp}

\subsection{Finetuning}
\label{sec:finetune}
The question instruction template with textual descriptions effectively translates cross-domain image classification tasks into text classification tasks. 
Zero-Shot LLM (ZS-LLM)~\cite{wei2021finetuned} feeds $\bT(\vx)$ into LLM and computes the sum of per-token log-likelihood for each category to apply \textit{rank classification} \cite{brown2020language}.
The category with the highest log-likelihood is chosen as the predicted category.
However, as shown in Figure~\ref{fig:train_pacs_all}, ZS-LLM performs poor in cross-domain tasks because
(1) As shown in Table \ref{tbl:word_frequency}, the distribution of textual description is still different across domains.  
LLM are susceptible to the influence of irrelevant context (\textit{e.g.,} domain-specific words) within the descriptions, which can negatively impact their classification accuracy \cite{shi2023large}.
(2) The output of LLM is correlated to but still misaligned with the ground-truth category \cite{zhong2021adapting}. See Appendix \ref{appendix:unpredictable_content} for further discussion.

To address the above issues and further exploit the capabilities of LLM in classifying samples based on textual descriptions,
we formulate the classification problem as a question-answer problem $(\bT(\vx), y)$,
where 
$y$, the class label of $\vx$, is represented by a sequence of class tokens and denotes the true answer to the question $\bT(\vx)$.

\subsubsection{Finetuning for DG Tasks}

Under the setting of DG, inspired by~\cite{cha2021swad, foret2020sharpness}, we assume that there exists a global labeling function that generates labels for multiple domains. Thus we combine the training datasets of all the source domains into a training dataset $\mathcal{D}$.
Then, instruction tuning is conducted on $\{(\bT(\vx), y): (\vx, y) \in \hD\}$ with the next-token prediction objective.
As LLMs have billions of parameters (\textit{e.g.}, LLaMA2 series have 7B, 13B, and 70B parameters),
fully finetuning the LLM is computationally expensive.
Therefore, we adopt the widely used parameter-efficient finetuning strategy (\textit{i.e.}, LoRA~\cite{hu2022lora}).
Let $\vtheta$ denote parameters in the LoRA modules.
Formally, the training objective of DG is to minimize the training loss as
\begin{equation}
\label{eqn:finetune}
\mathcal{L}_{dg} (\vtheta) = -  \sum_{(\vx,y) \in \mathcal{D}} \!\!\!\! \log p_{\vtheta} \big(y|\bT(\vx)\big).
\end{equation}

\subsubsection{Finetuning for UDA Tasks}
\label{sec:finetune_uda}
To make VLLaVO more suitable for UDA tasks, we propose the utilization of VLLaVO in conjunction with pseudo-labeling techniques.
Pseudo-labeling strategies \cite{sohn2020fixmatch, zhang2021flexmatch} have enjoyed success in UDA by leveraging the unlabeled target domain data \cite{lai2023padclip, venkateswara2017deep}.
Under the UDA setting, VLLaVO combines target domain samples with pseudo-labels and source domain samples with ground truth labels to finetune the LLM. 
Similar to Eq.~\eqref{eqn:finetune}, the pseudo-labeling-based finetuning objective of UDA is to minimize the training loss as 
\begin{align}
\label{eqn:retrain}
   \!\!\!\! \mathcal{L}_{uda}(\vtheta) \!= \! - \!\!\!\!\sum_{(\vx, y) \in \mathcal{D}}\!\!\!\!  \log  p_{\vtheta} \big( y|\bT(\vx)\big)  
    \! - \!\!\sum_{\vx \in \mathcal{T}} \log p_{\vtheta} \big( \hat{y}|\bT(\vx)\big),\!\!
\end{align}
where $\hat y$ denotes the pseudo-label for each target domain sample. 
Specifically, the LLM is first finetuned using the source domain training data $\mathcal{D}$ with a loss function identical to Eq.~\eqref{eqn:finetune}. Subsequently, the finetuned LLM is employed to generate pseudo-labels for target domain samples.

\subsection{Inference}
After finetuning LLM with the designed question instruction template, the LLM can make predictions from the category set. 
Thus, we use the finetuned LLM to evaluate the samples in the target domain $\mathcal{T}$ for both DG and UDA settings. 
Specifically, given a testing sample $\vx$, we feed the wrapped question $\bT(\vx)$ to the finetuned LLM to generate its answer as the prediction.

\section{Experiments}
In this section, we empirically evaluate the proposed VLLaVO method on both DG and UDA tasks.
Section \ref{sec:DG} discusses the experiments on DG tasks, and Section \ref{sec:UDA} discusses the experiments on UDA tasks. Additionally, to delve deeper into the capabilities of VLLaVO, we perform an ablation study in Section \ref{sec:ablation} and provide a detailed analysis in Section \ref{sec:analysis}.

\subsection{Experiments on DG Tasks}
\label{sec:DG}

\begin{table}[!t]\small
\centering
\caption{Testing accuracies (\%) of DG tasks on the \textit{PACS}, \textit{OfficeHome}, and \textit{DomainNet} datasets. n/a denotes not applicable. The best is in \textbf{bold}.}
\label{tbl:dg_pacs_officehome}
\vskip -.05in
\resizebox{\columnwidth}{!}{
\begin{tabular}{lccccc | ccccc | cccccc c}
    \toprule & \multicolumn{5}{c|}{\textit{PACS}}& \multicolumn{5}{c|}{\textit{OfficeHome}} & \multicolumn{7}{c}{\textit{DomainNet}} \\
    \midrule
    & A & C & P & S & \textbf{Avg} & A & C & P & R & \textbf{Avg} & C & I & P & Q & R & S & \textbf{Avg} \\
    \midrule
    ERM-ResNet~\cite{vapnik1999nature}  & 84.7 & 80.8 & 97.2 & 79.3 & 85.5 & 61.3 & 52.4 & 75.8 & 76.6 & 66.5 & 58.1 & 18.8 & 46.7 & 12.2 & 59.6 & 49.8 & 40.9  \\
    SWAD~\cite{cha2021swad} & 89.3 & 83.4 & 97.3 & 82.5 & 88.1 & 66.1 & 57.7 & 78.4 & 80.2 & 70.6 & 66.0 & 22.4 & 53.5 & 16.1 & 65.8 & 55.5 & 46.5  \\
    SAGM \cite{wang2023sharpness} & 87.4 & 80.2 & 98.0 & 80.8 & 86.6 & 65.4 & 57.0 & 78.0 & 80.0 & 70.1 & 64.9 & 21.1 & 51.5 & 14.8 & 64.1 & 53.6 & 45.0 \\
    MIRO \cite{cha2022domain} & 96.3 & 97.3 & 99.7 & 93.9 & 96.8 & 82.4 & 72.2 & 89.1 & 89.3 & 83.3 & 78.9 & 46.3 & 68.9 & 19.3 & 80.8 & 69.7 & 60.7  \\
    \midrule
    ZS-CLIP \cite{radford2021learning} & 96.4 & 98.9 & 99.9 & 87.7 & 95.7 & 80.7 & 64.6 & 86.3 & 88.0 & 79.9 & 70.7 & 49.1 & 66.4 & 14.8 & 82.7 & 63.1 & 57.8  \\
    ERM-CLIP \cite{vapnik1999nature} & 95.8 & 97.2 & 99.6 & 88.4 & 95.3 & 80.0 & 67.6 & 85.8 & 87.6 & 80.3 & 75.1 & 44.7 & 68.1 & 18.4 & 75.7 & 66.2 & 58.0\\
    DPL \cite{zhang2023domain} &-&-&-&-&97.3 &-&-&-&-&84.2 &-&-&-&-&-&-&56.7\\
    CLIPood \cite{shu2023clipood} & 98.9 & 99.6 & \textbf{100.0} & 91.2 & 97.4 & 88.0 & 74.0 & 92.8 &	93.2 & 87.0 & 77.6 & 54.7 & 72.5 & 20.7 & 85.7 & 69.9 & 63.5 \\
    \midrule
    ZS-LLM \cite{wei2021finetuned} & 87.7 & 94.1 & 97.4 & 93.8 & 93.3 & 59.0 & 53.0 & 70.0 & 67.0 & 62.2 & n/a &n/a&n/a&n/a&n/a&n/a&n/a\\
    \rowcolor{Gray}
    VLLaVO & \textbf{99.5} & \textbf{99.7} & \textbf{100.0} & \textbf{94.8} & \textbf{98.5} & \textbf{90.0} & \textbf{86.7} & \textbf{97.4} & \textbf{95.1} & \textbf{92.3} & \textbf{83.7} & \textbf{56.7} & \textbf{75.1} & \textbf{21.3} & \textbf{87.8} & \textbf{74.4} & \textbf{66.5} \\
    \bottomrule
\end{tabular}}
\vskip -0.2in
\end{table}

\noindent \textbf{Datasets.}
Experiments are conducted on four DG benchmark datasets, including \textit{PACS}~\cite{li2017deeper},
\textit{OfficeHome}~\cite{venkateswara2017deep}, and \textit{DomainNet}~\cite{fang2013unbiased}.
Due to the page limit, the statistics of each dataset are offered in Appendix \ref{appendix:dataset}.
We follow the leave-one-out cross-evaluation protocol~\cite{gulrajani2021in}: for each task, one domain is chosen as the target domain for testing, while the remaining domains serve as source domains for training.

\noindent \textbf{Baselines.} The proposed method is compared with existing DG methods including 
(i) ERM-ResNet \cite{vapnik1999nature}, SWAD \cite{cha2021swad}, MIRO \cite{cha2022domain}, and SAGM \cite{wang2023sharpness} that use only the image modality. 
(ii) Zero-shot CLIP (ZS-CLIP) \cite{radford2021learning}, ERM-CLIP, DPL \cite{zhang2023domain}, and CLIPood \cite{shu2023clipood} that use both the image and text modalities. 
ZS-CLIP uses the pre-trained CLIP to calculate the similarity between the query image embedding and the text embedding from the category set, and then choose the highest one as the prediction. 
For ERM-CLIP, following \cite{wortsman2022robust}, we add a classification head after the image encoder and finetune both components in the source domain.
(iii) ZS-LLM \cite{wei2021finetuned} classifies images based on the textual descriptions.
ZS-LLM and VLLaVO use the same VLMs 
(\textit{i.e.}, CLIP-ViT-H-14~\cite{radford2021learning} and BLIP~\cite{li2022blip})
for extracting textual descriptions and LLM (\textit{i.e.}, LLaMA-7B~\cite{touvron2023llama}) for classification. 
Note that the inference time of ZS-LLM grows with the number of categories, making it computationally infeasible for huge datasets like \textit{DomainNet} (with 345 categories). Therefore, ZS-LLM is not applicable \textit{DomainNet}.
Implementation details are provided in Appendix \ref{appendix:detail}.

\noindent
\textbf{Results.}
Table \ref{tbl:dg_pacs_officehome} shows the classification accuracies in the target domain for the \textit{PACS}, \textit{OfficeHome}, and \textit{DomainNet} datasets, respectively.
As can be seen, VLLaVO consistently achieves the best accuracy for each task on all the datasets.
VLLaVO outperforms CLIP-based methods (\textit{i.e.}, ZS-CLIP, ERM-CLIP, DPL, and CLIPood), showing that using LLM could help VLMs in learning domain-invariant features to boost cross-domain performance.
VLLaVO also outperforms ZS-LLM by a large margin (\textit{i.e.}, \textit{PACS}(+5.2\%), \textit{OfficeHome}(+30.1\%)), showing that finetuning the LLM on the designed question instruction template can improve its ability in handling vision cross-domain problems.
Those results demonstrate the effectiveness of the proposed VLLaVO method.

\begin{table}[!t]\small
\centering
\caption{Testing accuracies (\%) of UDA tasks on the \textit{OfficeHome} dataset. The best is in \textbf{bold}.}
\vskip -.05in
\label{tbl:uda_officehome}
\resizebox{\columnwidth}{!}{
\begin{tabular}{lcccccccccccc  c}
    \toprule  & A$\rightarrow$C  & A$\rightarrow$P  & A$\rightarrow$R & C$\rightarrow$A & C$\rightarrow$P & C$\rightarrow$R & P$\rightarrow$A & P$\rightarrow$C & P$\rightarrow$R & R$\rightarrow$A & R$\rightarrow$C & R$\rightarrow$P & \textbf{Avg} \\
    \midrule
    ERM \cite{vapnik1999nature}  & 44.1 & 67.1 & 74.3 & 53.3 & 62.0 & 64.5 & 51.9 & 38.9 & 72.9 & 64.5 & 43.8 & 75.4 & 59.4 \\
    DANN \cite{ganin2016domain} & 52.5 & 62.6 & 73.2 & 56.9 & 67.0 & 68.3 & 58.4 & 54.1 & 78.3 & 70.8 & 60.8 & 80.6 & 65.3   \\
    AFN \cite{AFN} & 52.6 & 72.4 & 77.0 & 64.9 & 71.1 & 72.9 & 64.1 & 51.3 & 77.8 & 72.2 & 57.5 & 82.1 & 68.0 \\
    CDAN \cite{long2018conditional} & 54.2 & 72.2 & 78.3 & 62.0 & 71.4 & 72.4 & 63.0 & 55.7 & 80.7 & 74.7 & 61.2 & 83.7 & 69.1 \\
    SDAT \cite{rangwani2022closer} & 58.2 & 77.5 & 81.4 & 66.1 & 76.5 & 76.4 & 63.7 & 56.7 & 82.5 & 76.0 & 62.1 & 85.2 & 71.9 \\
    MCC \cite{jin2020minimum}  & 56.8 & 79.8 & 82.7 & 67.8 & 77.0 & 77.8 & 67.0 & 55.4 & 81.8 & 74.0 & 61.4 & 85.4 & 72.2 \\
    ELS \cite{zhang2023free}  & 57.8 & 77.7 & 81.6 & 66.6 & 76.7 & 76.4 & 62.7 & 56.7 & 82.1 & 75.6 & 62.9 & 85.4 & 71.8 \\
    CDTrans \cite{xu2022cdtrans} & 68.8 & 85.0 & 86.9 & 81.5 & 87.1 & 87.3 & 79.6 & 63.3 & 88.2 & 82.0 & 66.0 & 90.6 & 80.5 \\
    PMTrans \cite{zhu2023patch} & 81.3 & 92.9 & 92.8 & 88.4 & 93.4 & 93.2 & \textbf{87.9} & 80.4 & 93.0 & 89.0 & 80.9 & 94.8 & 89.0 \\
    \midrule
    PADCLIP \cite{lai2023padclip} &76.4 &90.6 &90.8 &86.7 &92.3 &92.0 &86.0 &74.5 &91.5 &86.9 &79.1 &93.1& 86.7\\
    \rowcolor{Gray}
    VLLaVO & \textbf{85.4}	& \textbf{96.6} & \textbf{94.1} & \textbf{90.3} & \textbf{97.1} & \textbf{94.4} & \textbf{87.9} & \textbf{85.7} & \textbf{94.5} &	\textbf{90.1} &	\textbf{85.5}&	\textbf{97.3} & \textbf{91.6}\\
    \bottomrule
\end{tabular}
}
\vskip -0.1in
\end{table}

\subsection{Experiments on UDA Tasks}
\label{sec:UDA}

\noindent
\textbf{Datasets.}
For UDA,
experiments are conducted on two benchmark datasets: 
\textit{OfficeHome}~\cite{venkateswara2017deep} and \textit{DomainNet}~\cite{fang2013unbiased}.
The statistics of those datasets are provided in Table~\ref{table:ds_UDA} of Appendix \ref{appendix:dataset}.

\noindent
\textbf{Baselines.}
The proposed method is compared with various UDA
methods using the image modality only, including ERM \cite{vapnik1999nature}, DANN \cite{ganin2016domain}, AFN \cite{AFN}, CDAN \cite{long2018conditional}, SDAT \cite{rangwani2022closer}, MCC \cite{jin2020minimum}, ELS \cite{zhang2023free}, CDTrans \cite{xu2022cdtrans}, and PMTrans \cite{zhu2023patch}. We also compare with PADCLIP \cite{lai2023padclip} that uses both the image and text modalities. 
Implementation details are provided in Appendix \ref{appendix:detail}.

\noindent
\textbf{Results.}
Tables \ref{tbl:uda_officehome} and \ref{tbl:uda_domainnet}  
show the testing accuracies of UDA tasks on the \textit{OfficeHome} and \textit{DomainNet} dataset, 
respectively. 
As can be seen, VLLaVO consistently achieves superior average accuracy on all datasets.
Compared with baselines based on the image modality only (\textit{i.e.}, ERM, DANN, AFN, CDAN, SDAT, MCC, ELS, CDTrans, and PMTrans), VLLaVO performs better on average, demonstrating the effectiveness of using both the image and text modalities. 
VLLaVO also has good performance for each task on all the datasets.
Specifically, VLLaVO achieves the best performance on all 12 tasks and 23 of 30 tasks on the \textit{OfficeHome} and \textit{DomainNet} 
datasets, respectively.
Moreover, VLLaVO outperforms the CLIP-based method (\textit{i.e.}, PADCLIP), indicating that incorporating LLMs into cross-domain learners is more effective in cross-domain learning.

\begin{table*}[!t]\footnotesize
\centering
\caption{Testing accuracies (\%) of UDA tasks on the \textit{DomainNet} dataset.}
    \vskip -.1in
    \label{tbl:uda_domainnet}
    \setlength{\tabcolsep}{0.7mm}{
    \resizebox{\columnwidth}{!}{
    \begin{tabular}{c|ccccccc|cccccccc|cccccccc}
    \toprule ERM & C  & I  & P & Q & R & S & \textbf{Avg} & MCC & C & I & P & Q & R & S & \textbf{Avg} & CDTrans & C & I & P & Q & R & S & \textbf{Avg} \\
    \midrule
    \midrule
    C  & - & 18.0 & 36.2 & 12.1 & 53.6 & 42.6 & 32.5 & C & - & 19.9 & 39.6 & 9.0 & 56.9 & 43.6 & 33.8 & C & - & 29.4& 57.2& 26.0& 72.6 & 58.1 &48.7 \\
    I  & 41.0 & - & 35.7 & 4.7 & 52.9 & 31.2 & 33.1 & I & 37.2 & - & 38.1 & 3.0 & 54.8 & 26.6 & 31.9 & I & 57.0 & - & 54.4 & 12.8 & 69.5 & 48.4 & 48.4  \\
    P  & 46.0 & 18.5 & - & 6.2 & 59.9 & 38.5 & 34.4 & P & 48.4 & 19.7 & - & 4.4 &61.1 & 41.2 &35.0&P&62.9&27.4&-&15.8&72.1&53.9&46.4 \\
    Q  & 13.6 & 1.2 & 2.1 & - & 6.0 & 10.2 & 6.6  & Q  & 18.5  &3.9  & 9.2 &  - & 17.6 &13.0&12.4& Q &44.6&8.9&29.0&-&42.6&28.5&30.7\\
    R  & 51.8 & 21.8 & 50.3  & 7.4  & -  & 38.0  & 33.9  & R & 55.1 & 22.5  &54.0   & 4.7  & - &37.8&34.8&R&66.2&31.0&61.5&16.2&-&52.9&45.6\\
    S   &  54.8 & 15.6  & 38.6  & 14.1  & 49.9  & -  & 34.6  &  S & 60.0  &  18.7 &  47.3 &  10.3 & 57.9 &-&38.8&S&69.0&29.6&59.0&27.2&72.5&-&51.5\\
    \textbf{Avg}  &42.0  & 15.0 &  32.6& 8.9 & 44.5 & 32.1 & 29.2 & \textbf{Avg} & 43.8 & 16.9 & 37.6 & 6.3 & 49.6 & 32.4 & 32.4 & \textbf{Avg} & 59.9 & 25.3 & 52.2 & 19.6 & 65.9 & 48.4 & 45.2\\
    \midrule
    \midrule
    PMTrans  & C  & I  & P & Q & R & S & \textbf{Avg} & PADCLIP & C  & I  & P & Q & R & S & \textbf{Avg} & VLLaVO & C & I & P & Q & R & S & \textbf{Avg}\\
    \midrule
    \midrule
    C & - & 34.2 & 62.7 & 32.5 & 79.3 & 63.7 & 54.5 & C & - & 55.1 & 71.1 & 36.8 & 84.2 & 68.1 & 63.1 & C & - & 56.0 & 71.5 & 19.9 & 87.1 & 74.2 & 61.8 \\
    I & 67.4 & - & 61.1 & 22.2 & 78.0 & 57.6 & 57.3 & I & 73.6 & - & 70.6 & 18.0 & 83.5 & 66.6 & 62.5 & I & 82.0 & - & 72.7 & 21.5 & 86.9 & 72.9 & 67.2 \\
    P  & 69.7 & 33.5 & - & 23.9 & 79.8 & 61.2 & 53.6 & P &  75.4&  54.3& - & 32.0 & 83.5 & 67.2 & 62.5 & P & 82.3 & 55.3 & - & 17.5 & 86.8 & 72.8 & 63.0 \\
    Q & 54.6 & 17.4 & 38.9 & - & 49.5 & 41.0 & 40.3 & Q & 74.6 &  53.6 &  70.0& - & 83.1 & 66.1 & 69.5 & Q & 79.2 & 52.5 & 66.3 & - & 84.8 & 70.7 & 70.7 \\
    R  &  74.1&  35.3&  70.0&  25.4& - & 61.1 & 53.2 & R  &  76.4&  54.9&  72.7&  31.7& - & 67.5 & 60.6 & R & 83.7 & 55.7 & 74.2 &18.8&-&73.6&61.2\\
    S  & 73.8 & 33.0&  62.6&  30.9&  77.5& - & 55.6 & S  &  76.3&  54.9&  71.7&  34.9&  83.6& - & 64.3 & S & 83.1 & 55.3 & 75.4 & 21.4 & 87.1 & - & 64.4\\
    \textbf{Avg}  & 67.9 & 30.7 & 59.1 & 27.0 & 72.8 & 56.9 & 62.9 & \textbf{Avg}  & 75.3 & 54.6 & 71.2 & 30.7 &  83.6& 67.1 & 63.7 & \textbf{Avg} & 82.1 & 55.0 & 72.0 & 19.8 & 84.2 & 72.8 & \textbf{64.7}\\
    \bottomrule
\end{tabular}}}
\vskip -0.1in
\end{table*}

\subsection{Ablation Study}
\label{sec:ablation}

\begin{table}[!t]\footnotesize
\centering
\caption{Effects of different textual information for DG tasks on the \textit{OfficeHome} dataset. The best is in \textbf{bold}.}
\label{tbl:ablation_component_DG}
\vskip -.1in
\begin{NiceTabular}{c@{\hskip7pt}c@{\hskip7pt}c|c@{\hskip7pt}c@{\hskip7pt}c@{\hskip7pt}c@{\hskip7pt}c}
    \toprule 
    Tags & Attributes & Captions & A & C & P & R & \textbf{Avg} \\
    \midrule
    \cmark & \xmark &\xmark & 89.2&	\textbf{85.9} &	97.3&	94.7& 91.8 \\
    \xmark & \cmark & \xmark &88.9&	\textbf{85.9} &	96.9&	94.4& 91.5 \\
    \xmark & \xmark & \cmark &88.2 & 83.3 & 95.3 & 92.8 & 89.9 \\
    \cmark & \cmark &\cmark  & \textbf{90.5} & 85.8 & \textbf{97.4} & \textbf{95.4} & \textbf{92.3} \\
    \midrule
    \multicolumn{3}{c}{Previous SOTA\cite{shu2023clipood}} & 88.0 & 74.0 & 92.8 & 93.2 & 87.0 \\
    \bottomrule
\end{NiceTabular}
\vskip -.25in
\end{table}

\subsubsection{Effects of Description Components}
\label{sec:ablation_VLM}
By following the setting in Section~\ref{sec:DG}, we conduct experiments to study the effects of tags, attributes, and captions in the descriptions extracted by VLMs on the performance.
Table \ref{tbl:ablation_component_DG} shows the testing accuracy of DG tasks on the \textit{OfficeHome} dataset.
As can be seen,
VLLaVO using only the tags, attributes or captions performs better than the existing SOTA method (\textit{i.e.}, CLIPood) shown in Table \ref{tbl:dg_pacs_officehome}. 
By using tags, attributes, and captions as descriptions, VLLaVO achieves the best performance, which demonstrates the effectiveness of using tags, attributes, and captions together.
Due to page limit, experimental results of UDA tasks on the \textit{OfficeHome} dataset are offered in Appendix \ref{appendix:result}. Again, VLLaVO using all the text descriptions achieves the best performance.

To study the effect of different VLMs/LLMs for generating the descriptions, we replace the default 
caption model (\textit{i.e.}, BLIP) with GIT\cite{wang2022git} and the default vocabulary generating model (\textit{i.e.}, GPT-3) with GPT-3.5, respectively.
Table \ref{tbl:ablation_description_models} shows the testing accuracy of DG tasks on the \textit{OfficeHome} dataset. As can be seen, all variants of VLLaVO outperform the previous SOTA.

\begin{table}[!t]\footnotesize
    \centering
    \caption{Effects of different models to generate descriptions for DG tasks on the \textit{OfficeHome} dataset. The best is in \textbf{bold}.}\vskip -.05in
    \label{tbl:ablation_description_models}
    \begin{tabular}{l@{\hskip7pt}c@{\hskip7pt}c@{\hskip7pt}c@{\hskip7pt}c@{\hskip7pt}c}
        \toprule   & A & C & P & R & \textbf{Avg} \\
        \midrule
        Previous SOTA\cite{shu2023clipood} &  88.0 & 74.0 & 92.8 & 93.2 & 87.0 \\
        VLLaVO (BLIP\;+\;GPT-3) & \textbf{90.0} & \textbf{86.7} & \textbf{97.4} & \textbf{95.1} & \textbf{92.3} \\ 
        VLLaVO (GIT\;\;\;+\;GPT-3)  & 87.8 & 82.0 & 96.6 & 94.7 & 90.3 \\
        VLLaVO (BLIP\;+\;GPT-3.5)  & 89.7 & 86.6 & 97.3 & \textbf{95.1} & 92.2 \\
        \bottomrule
    \end{tabular}
    \vspace{-0.1in} 
\end{table}

\begin{table}[!t]\footnotesize
\centering
\caption{Effects of different LLM components for DG tasks on the \textit{OfficeHome} dataset. The best is in \textbf{bold}.}
\label{tbl:ablation_llm_dg}
\vskip -.05in
\begin{tabular}{lccccc}
\toprule
 & A & C & P & R & \textbf{Avg} \\
\midrule
Previous SOTA\cite{shu2023clipood} & 88.0 & 74.0 & 92.8 & 93.2 & 87.0 \\
VLLaVO (w/ Flan-T5-base) & 85.0 & 81.7 & 92.8 & 90.4 & 87.5 \\
VLLaVO (w/ LLaMA2-7B) & \textbf{90.0} & \textbf{86.7} & \textbf{97.4} & \textbf{95.1} & \textbf{92.3} \\
\bottomrule
\end{tabular}
\vspace{-0.25in} 
\end{table}

\subsubsection{Effects of LLM}
To verify the effectiveness of LLMs used in VLLaVO, we replace the default LLaMA2-7B with Flan-T5-base~\cite{chung2022scaling}, 
a language model with a different structure and much fewer parameters (\textit{i.e.}, 250 million).
Table \ref{tbl:ablation_llm_dg} shows the testing accuracy of DG tasks on the \textit{OfficeHome} dataset by using the setting in Section~\ref{sec:DG}.  
As can be seen, VLLaVO (w/ Flan-T5-base) performs slightly better than the previous SOTA (\textit{i.e.}, CLIPood),
showing that it is flexible to choose the LLM for VLLaVO.
Moreover, using a more powerful LLM for VLLaVO (\textit{e.g.}, LLaMA2-7B) yields better performance.
Table \ref{tbl:ablation_llm_uda} of Appendix \ref{appendix:result} shows the testing accuracy of UDA tasks on the \textit{OfficeHome} dataset.
Again, both VLLaVO (w/ Flan-T5-base) and VLLaVO (w/ LLaMA2-7B) perform better than the previous SOTA.


\subsubsection{Effects of Finetuning with Pseudo-Labels in UDA}
\begin{wraptable}{r}{.4\linewidth}
    \vskip -.5in
    \centering
    \caption{Testing Accuracy of VLLaVO (w/o PseudoLabels) for UDA tasks on the \textit{DomainNet} dataset.}
    \vskip -.02in
    \label{tbl:uda_domainnet_ablation}
    \resizebox{1.0\linewidth}{!}{
    \begin{tabular}{c|ccccccc}
    \toprule  & C & I & P & Q & R & S & \textbf{Avg} \\
    \midrule
    C & - & 54.9 & 70.7 & 19.9 & 86.5 & 73.9& 61.2 \\
    I  & 81.5 & - & 72.2 & 21.3 & 86.6 & 72.9 & 66.9  \\
    P  & 81.4 & 54.5 & - & 17.5 & 86.3 & 72.2 &62.4 \\
    Q  & 77.2 & 49.6 & 64.5 & - & 82.9 & 69.3 & 68.7\\
    R  & 82.8 & 55.0 & 72.6 & 18.6 &- & 73.2 &60.4 \\
    S   & 82.2 & 54.2 &74.5  &21.4 & 86.3& - & 63.7\\
    \textbf{Avg} & 81.0 & 53.6 & 70.9 & 19.7 & 85.7 & 72.3 & 63.9\\
    \bottomrule
\end{tabular}}
\vskip -0.7in
\end{wraptable}
To verify the effectiveness of finetuning with pseudo-labels in UDA, we compare VLLaVO with its variant, VLLaVO (w/o PseudoLabels), on the \textit{DomainNet} dataset.
Here, VLLaVO (w/o PseudoLabels) represents VLLaVO without the pseudo-labeling-based finetuning in Section \ref{sec:finetune_uda} and it uses Eq. \eqref{eqn:finetune} as the training loss. 
As shown in Table \ref{tbl:uda_domainnet_ablation}, VLLaVO (w/o PseudoLabels) is worse than VLLaVO (Table \ref{tbl:uda_domainnet}), showing that finetuning with pseudo-labels can boost the performance of VLLaVO. 
We further investigate the impact of the number of finetuning times on performance.
Specifically, for each time of finetuning, we utilize the pseudo-label generated by the previous finetuned LLM. As shown in Table \ref{tbl:uda_officehome_retrain} of Appendix \ref{appendix:result}, increasing the number of finetuning boosts the performance and will converge to the maximum. 
Considering the efficiency, our default VLLaVO entails a single pseudo-labeling-based finetuning for UDA.

\subsection{Analysis}
\label{sec:analysis}
In this section, we aim to answer the following questions: 
(1) How can VLLaVO alleviate domain shifts?
(2) How about the zero-shot capability of VLLaVO?
(3) What information does the LLM within VLLaVO focus on?

\subsubsection{VLLaVO is a Good Domain-Invariant Feature Learner.}
\label{sec:domain_invariant}

Figure~\ref{fig:tsne_demo} shows the t-SNE visualization~\cite{van2008visualizing} of feature embeddings of samples extracted from two domains (\textit{i.e.}, Art painting and Cartoon) on the \textit{PACS} dataset, where CLIP embeddings are extracted from the penultimate layer of the pre-trained CLIP image encoder,
while LLM embeddings are obtained by averaging the last hidden state of the pre-trained LLaMA with $\bT(\vx)$ as the input.
As can be seen, in the visual modality shown in Figure~\ref{fig:tsne_image_demo},
samples from the same category have a noticeable domain shift, while, in the textual modality as shown in Figure~\ref{fig:tsne_text_demo}, samples from the same category are very close across the two domains, showing that the LLM can alleviate domain shifts.
This observation gains additional support from  Figure~\ref{fig:tsne_officehome} in Appendix~\ref{appendix:result}, which presents the t-SNE visualization of Art and Real World domains on the \textit{OfficeHome} dataset.
\begin{wraptable}{r}{.35\linewidth}
\centering
\vskip -.25in
\caption{Word frequency per sample on the \textit{PACS} dataset}
\label{tbl:word_frequency}
\resizebox{1.0\linewidth}{!}{
\begin{tabular}{c@{\hskip3pt}c@{\hskip6pt}c@{\hskip6pt}c@{\hskip6pt}c}
    \toprule words & A & C & P & S \\
    \midrule
    painting & \textbf{6.72} & 0.03 & 0.03 & 0.01 \\
    cartoon & 0.12 & \textbf{3.77} & 0.00 & 0.36 \\
    camera & 0.12 & 0.12 & \textbf{0.57} & 0.06 \\
    drawing & 0.56 & 0.28 & 0.02 & \textbf{4.22} \\
    \bottomrule
\end{tabular}
} \vskip -.2in
\end{wraptable}

Moreover, Table \ref{tbl:word_frequency} shows the frequency of domain-specific words per sample on the \textit{PACS} dataset.
As can be seen, the distribution of domain-specific words is different across domains, demonstrating that a domain gap exists in textual descriptions. 
Despite this domain gap, the finetuned LLMs exhibit strong performance, thereby confirming the effective handling of the visual domain gap by LLMs.
Those results explain why the proposed VLLaVO uses LLMs to handle domain shifts in cross-domain learning. 

\subsubsection{VLLaVO is a Good Zero-Shot Learner.}
\label{sec:llavo_zero_shot}
To study the zero-shot ability of VLLaVO,
we evaluate the cross-dataset performance of VLLaVO and zero-shot LLM (ZS-LLM) on DG datasets.
Specifically, we select a DG dataset with the minimum number of samples (\textit{i.e.}, \textit{PACS}) as the source dataset, while considering the remaining DG datasets as the target datasets.
All domains in the source dataset are merged for finetuning, and all domains in each target dataset are merged as well for testing. 
In this case, the cross-dataset setting is to test the zero-shot learning ability.
For ZS-LLM, we follow \cite{wei2021finetuned} and directly feed $\bT(\vx)$ into LLMs and compare the log-likelihood of each category. The category with the highest log-likelihood is chosen as the prediction.
As mentioned in Section \ref{sec:DG}, the inference time of ZS-LLM is very costly for the \textit{DomainNet} dataset with 345 categories.
To reduce the computational demands associated with applying ZS-LLM to the \textit{DomainNet} dataset, we create a subset of \textit{DomainNet}, referred to as \textit{subDomainNet}. This subset consists of the first 32 categories, arranged in alphabetical order.

\begin{figure}[!t]
\centering
\!\!\!\!\!\!
\subcaptionbox{
Accuracy.
\label{fig:train_pacs_all}
}
{\includegraphics[width=0.4\textwidth]{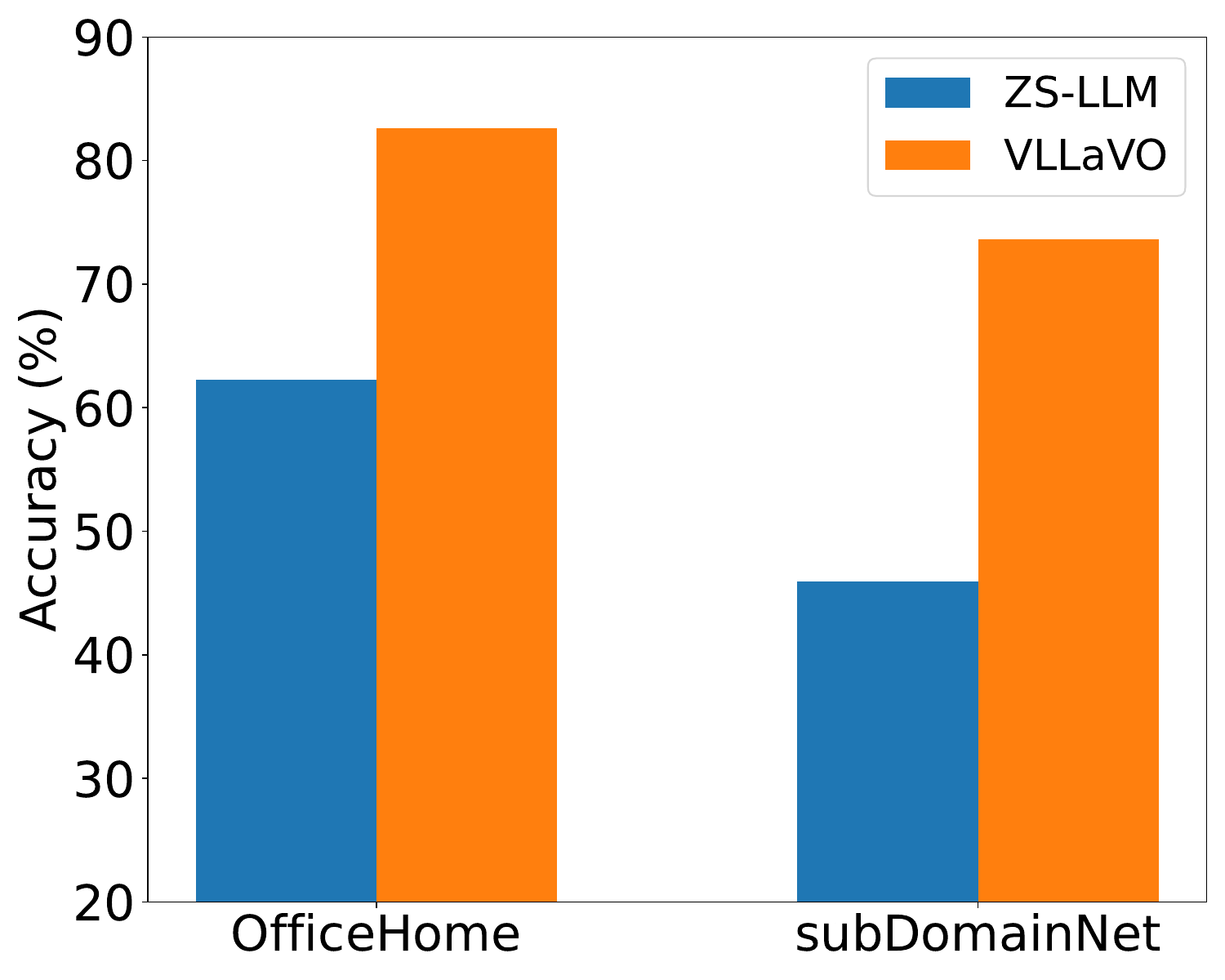}}
\subcaptionbox{
Inference time per sample.
\label{fig:time_complexity}
} {\includegraphics[width=0.4\textwidth]{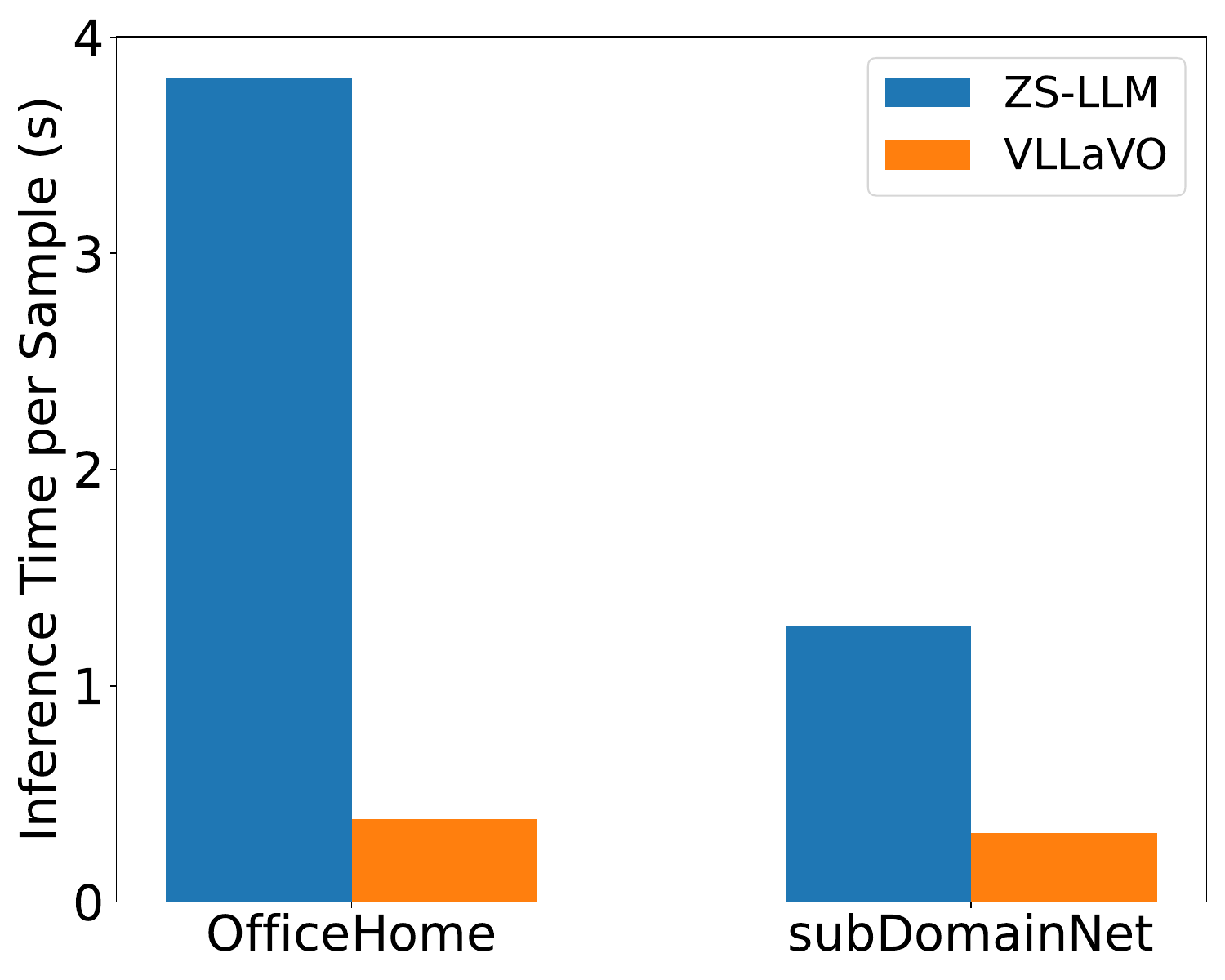}}
\!\!\!\!\!\!
\caption{Comparison between ZS-LLM and VLLaVO (finetuned on the \textit{PACS} dataset).
}
\label{fig:histogram}
\vspace{-0.3in} 
\end{figure}

Figure~\ref{fig:train_pacs_all} shows the accuracy comparison between ZS-LLM and VLLaVO.
As can be seen, the proposed VLLaVO outperforms ZS-LLM on the \textit{OfficeHome} and \textit{subDomainNet} datasets, verifying the superior zero-shot capability of VLLaVO.
This can be attributed to the finetuning of VLLaVO on the designed question instruction template on \textit{PACS}.
This process significantly enhances the LLM understanding of the directives for the template $\bT$. It enables more accurate predictions by deepening the comprehension of the relationships between $\mathbb{D}(\vx)$ and the category set.

To verify the efficiency of VLLaVO under the zero-shot condition, we compare the inference time per sample between VLLaVO and ZS-LLM. As shown in Figure~\ref{fig:time_complexity}, compared with ZS-LLM, the inference of VLLaVO is consistent much faster on the \textit{OfficeHome} and \textit{subDomainNet} datasets.

\subsubsection{What Information Does VLLaVO Focus On?}
\label{sec:focus}

To determine the crucial information that VLLaVO prioritizes, we use the gradient of the objective function with respect to each token embedding as a measure of sensitivity and importance, as discussed in \cite{he2023sensitivity,feng2018pathologies}, to select important words within the descriptions. 
Specifically, 
for each word in the description, we calculate the mean absolute value of all the token gradients associated with that word as their sensitivity score. We then identify the Top-10 words with the highest scores by excluding the stop-words.

Example \ref{ex:exmp1} shows an image of a house from the cartoon domain on the \textit{PACS} dataset, and words marked by the red represent the Top-10 sensitive words that the LLM in VLLaVO primarily focuses on. 
As can be seen, the sensitive words for the LLM can be categorized into two groups:
(1) category words such as ``house'' and ``bird'' in the category set and (2) words related to the ground-truth label, \eg., ``schoolhouse'', ``birdhouse'', ``playhouse'', and ``apartment''. 
Thus, after finetuning the LLM with the designed question instruction template, 
VLLaVO can focus on domain-invariant words that are related to classification tasks and disregard irrelevant words, further verifying the effectiveness of the proposed VLLaVO.
More examples are shown in Appendix \ref{appendix:description_examples}.
\begin{exmp}{Textual Description $\mathbb{D}(\vx)$ of Image.}{exmp1}
    \footnotesize
    \vspace{-0.03in}
    \begin{minipage}[]{0.80\textwidth}
    {\small \textbf{Tags:}} \\
    -\textcolor{red}{schoolhouse} \\
    -Property manager \\
    -\textcolor{red}{birdhouse} \\
    -Gingerbread \textcolor{red}{house} \\
    -\textcolor{red}{Playhouse} 
    \end{minipage}%
    \begin{minipage}[]{0.20\textwidth}
    \raggedleft
    \includegraphics[height=0.9in]{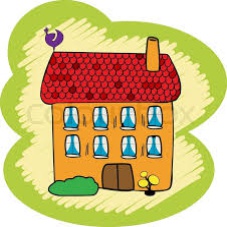}
    \vspace{-0.23in}
    \begin{center}
        (label:~house)
    \end{center}
    \end{minipage}
    {
    \small \textbf{Attributes}
    }: \\
    -\textcolor{red}{house} which is a building with walls and a roof \\
    -residence which has windows \\
    -property which has \textcolor{red}{house}, \textcolor{red}{apartment}, or other structure \\
    -\textcolor{red}{schoolhouse} which typically large and rectangular in shape \\
    -resident which has Background features such as furniture, buildings, other people \\ 
    
    {\small \textbf{Captions}}: \\
    -a picture of a \textcolor{red}{house} with a \textcolor{red}{bird} on the roof \\
    -a cartoon \textcolor{red}{house} with a lot of windows \\
    -a little \textcolor{red}{house} is outside a green bush \\
    -a \textcolor{red}{house} with a \textcolor{red}{bird} sitting on top of it \\
    -two story \textcolor{red}{house} with a roof with a brick chimney \\
    -a cartoon \textcolor{red}{house} with a small window and red hat \\
    -the cartoon \textcolor{red}{house} is next to a \textcolor{red}{bird} perched on a building \\
    -a red roof and a yellow \textcolor{red}{house} with a \textcolor{red}{bird} \\
    -a small brick building with a \textcolor{red}{bird} on top \\
    -a \textcolor{red}{house} with a \textcolor{red}{bird} on the roof
\end{exmp}

\section{Conclusion}
In this paper, we propose VLLaVO to address visual domain shifts in visual cross-domain learning.
VLLaVO uses VLMs for bridging the image and text modalities and finetunes the LLM using the designed question instruction template to focus on the domain-invariant features that are relevant to labels.
VLLaVO excels in accurately identifying the correct category. The VLMs and LLM within VLLaVO could be substituted with any large-scale pretrained models.
Extensive experiments demonstrate that VLLaVO achieves state-of-the-art performance in both DG and UDA benchmarks. 
To facilitate future research, we release a series of datasets in text modality for public use.
In our future work, we will extend VLLaVO to other learning tasks.

\bibliographystyle{splncs04}
\bibliography{main}
\clearpage

\begin{Large}
\begin{center}
\appendix{\textbf{Appendix}}
\end{center}
\end{Large}

\vspace{-0.05in}
\section{Experiment}
\label{appendix:experiment}
\subsection{Datasets.}
\label{appendix:dataset}
Tables~\ref{table:ds_DG} and~\ref{table:ds_UDA} summarize the number of images, classes, domains, and tasks of datasets under the DG and UDA settings, respectively.
Taking the \textit{PACS} dataset as an example, the set of categories is ``dog, elephant, giraffe, guitar, horse, house, person''.

\begin{table}[tbph]\small
    \vskip -0.3in
    \centering
    \caption{Statistics of the DG datasets used.} \label{table:ds_DG} 
    \vskip -.15in
    \begin{tabular}{c c c c c}
        \toprule
 Dataset        & \#images & \#classes & \#domains & \#tasks \\
        \midrule
        \textit{PACS}  & $9,991$ & $7$  & $4$ & $4$ \\
        \textit{OfficeHome}  & $15,500$ & $65$ & $4$ & $4$ \\
        \textit{DomainNet}  & $586,575$ & $345$  & $6$ & $6$ \\
        \bottomrule
    \end{tabular}
    \end{table}

\begin{table}[!tbph]\small
    \vskip -0.5in
    \centering
    \caption{Statistics of the UDA datasets used.}
    \label{table:ds_UDA}
    \vskip -.1in
    \begin{tabular}{c c c c c}
        \toprule
Dataset         & \#images & \#classes & \#domains & \#tasks \\
        \midrule
        \textit{OfficeHome}  & $15,500$ & $65$ & $4$ &  $12$ \\
        \textit{DomainNet}  & $586,575$ & $345$  & $6$ & $30$ \\
        \bottomrule
    \end{tabular}
    \vskip -0.5in
    \end{table}

\subsection{Implementation Details.}
\label{appendix:detail}
    
For the DG experiments, we utilize the CLIP-ViT-H-14 model\footnote{\url{https://huggingface.co/laion/CLIP-ViT-H-14-laion2B-s32B-b79K}} as the default CLIP model for extracting tags and attributes. For generating captions, we employ the BLIP-large captioning checkpoint\footnote{\url{https://huggingface.co/Salesforce/blip-image-captioning-large}}.
The values of $K$, $M$, and $N$ used in the image-to-text extraction are all set to $5$.
We adopt LLaMA-2-7B~\cite{touvron2023llama} as the base LLM, and train it by LoRA finetuning~\cite{hu2022lora} with rank $8$ and $\alpha$ as 16. 
We use a batch size of $128$ and the AdamW optimizer \cite{loshchilov2018decoupled} with a learning rate of $0.001$, $\beta_1=0.9$, $\beta_2=0.999$, and $\epsilon=10^{-8}$. 
The LLM is finetuned for $100$ steps on all datasets except the largest \textit{DomainNet} dataset which uses $200$ steps. 
To generate answers, we employ beam search~\cite{sutskever2014sequence} with a beam width of $4$.
    
For the UDA experiments, we finetune LLM and finetune LLM with pseudo-labels for both $2$ epochs. The other training settings are consistent with the above DG.
    
\subsection{More Experimental Results.}
\label{appendix:result}

Table \ref{tbl:ablation_component_UDA} shows the testing accuracies of VLLaVO with variants of the textual description for UDA tasks on the \textit{OfficeHome} dataset. As can be seen, VLLaVO with tags, attributes, and captions achieves the best performance.


Table \ref{tbl:ablation_llm_uda} shows the testing accuracies of different LLMs for UDA tasks on the \textit{OfficeHome} dataset. As can be seen, VLLaVO (w/ LLaMA2-7B) achieves the best performance, and VLLaVO (w/ FLAN-T5-base) still outperforms the previous SOTA method.

\begin{figure}[!t]
\centering
\!\!\!\!\!\!
\subcaptionbox{
CLIP embeddings.
\label{fig:tsne_image_officehome}
}
{\includegraphics[width=0.45\textwidth]{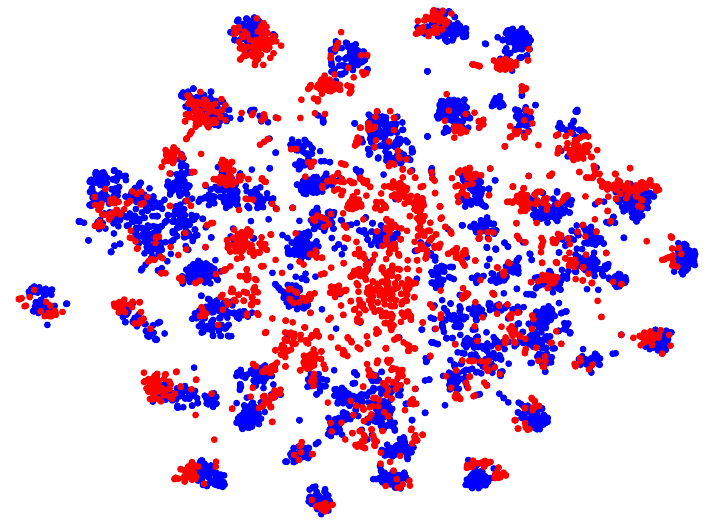}}
\quad\;
\subcaptionbox{
LLM embeddings.
\label{fig:tsne_text_officehome}
} {\includegraphics[width=0.45\textwidth]{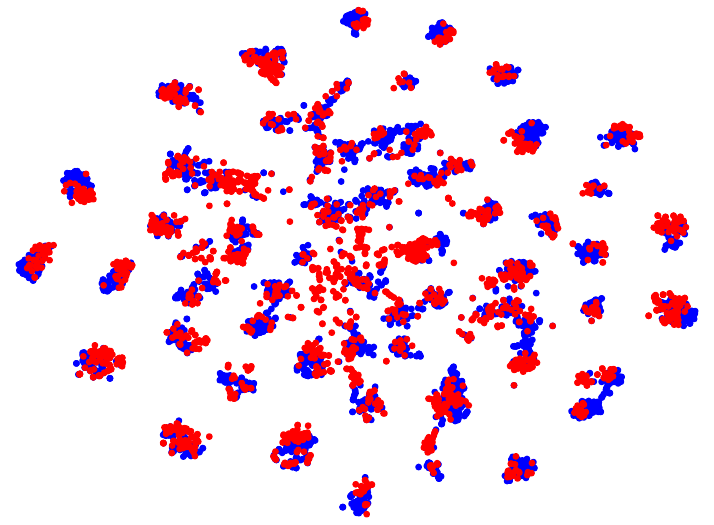}}
\!\!\!\!\!\!
\vskip -.1in
\caption{t-SNE visualization of samples from the Real World domain (marked in red color) and the Art domain (marked in blue color) on the \textit{OfficeHome} dataset.}
\label{fig:tsne_officehome}
\vspace{-0.4cm} 
\end{figure}

Figure \ref{fig:tsne_officehome} shows the t-SNE visualization of feature embeddings of samples extracted from two domains (\textit{i.e.}, Art and Real World) on the \textit{OfficeHome} dataset, where CLIP embeddings are extracted from the penultimate layer of pretrained CLIP image encoder, while LLM embeddings are obtained by averaging the last hidden state of pretrained LLaMA with $\bT(x)$ as the input. 
As can be seen, the domain shift in LLM embeddings is smaller than that of CLIP embeddings, indicating the effectiveness of using LLM embeddings in cross-domain learning.

\begin{table*}[!htbp]
\centering
\caption{Effects of different textual descriptions for UDA tasks on the \textit{OfficeHome} dataset. The best is in \textbf{bold}.}
\label{tbl:ablation_component_UDA}
\vskip -.15in
\resizebox{\columnwidth}{!}{
\begin{NiceTabular}{ccc|cccc cccc cccc  c}
    \toprule 
    Tags&Attributes & Captions & A$\rightarrow$C  & A$\rightarrow$P  & A$\rightarrow$R & C$\rightarrow$A & C$\rightarrow$P & C$\rightarrow$R & P$\rightarrow$A & P$\rightarrow$C & P$\rightarrow$R & R$\rightarrow$A & R$\rightarrow$C & R$\rightarrow$P & \textbf{Avg} \\
    \midrule
    \cmark & \xmark& \xmark & 77.3&	90.3	&89.2&	83.7&	94.0	&91.6&	82.3	&78.8&	90.4&	83.1&	79.5	&94.3	&86.2 \\
    \xmark & \cmark & \xmark& 76.5&	94.0&	89.7&	79.2&	95.4&	91.5&	78.0&	78.4&	90.5&	78.8&	78.2&	95.2&	85.4 \\
    \xmark & \xmark & \cmark& 85.1 & 95.5 & 91.3 &	87.8 &	94.9 & 91.7 & 84.8 & 83.2 &	90.9 &	87.8& 82.8 & 95.2 &	89.3 \\
    \cmark & \cmark  &\cmark  & \textbf{85.4}	& \textbf{96.6} & \textbf{94.1} & \textbf{90.3} & \textbf{97.1} & \textbf{94.4} & \textbf{87.9} & \textbf{85.7} & \textbf{94.5} &	\textbf{90.1} &	\textbf{85.5}&	\textbf{97.3} & \textbf{91.6} \\
    \bottomrule
\end{NiceTabular}}
\end{table*}
\vspace{-0.5cm} 


\begin{table*}[!htbp]\small
\centering
\caption{Effects of different LLMs for UDA tasks on the \textit{OfficeHome} dataset. The best is in \textbf{bold}.}
    \vskip -.1in
    \label{tbl:ablation_llm_uda}
    \resizebox{\columnwidth}{!}{
    \begin{tabular}{lcccccccccccc  c}
    \toprule Method & A$\rightarrow$C & A$\rightarrow$P & A$\rightarrow$R & C$\rightarrow$A & C$\rightarrow$P & C$\rightarrow$R & P$\rightarrow$A & P$\rightarrow$C & P$\rightarrow$R & R$\rightarrow$A & R$\rightarrow$C & R$\rightarrow$P & \textbf{Avg} \\
    \midrule
    Previous SOTA \cite{lai2023padclip} &76.4 &90.6 &90.8 &86.7 &92.3 &92.0 &86.0 &74.5 &91.5 &86.9 &79.1 &93.1& 86.7\\
    VLLaVO (w/ FLAN-T5-base) & 82.0 & 94.4 & 89.7 & 86.1 & 92.6 & 90.4 & 85.6 & 83.9 & 88.5 & 88.8 & 84.5 & 93.3 & 88.3 \\
    VLLaVO (w/ LLaMA2-7B) & \textbf{85.4}	& \textbf{96.6} & \textbf{94.1} & \textbf{90.3} & \textbf{97.1} & \textbf{94.4} & \textbf{87.3} & \textbf{85.7} & \textbf{94.5} &	\textbf{90.1} &	\textbf{85.5}&	\textbf{97.3} & \textbf{91.5}\\
    \bottomrule
\end{tabular}}
\end{table*}

\begin{table*}[!htbp]\small
\centering
\caption{Effects of the time of finetuning with pseudo-label for UDA tasks on the \textit{OfficeHome} dataset. The best is in \textbf{bold}.}
    \vskip -.1in
    \label{tbl:uda_officehome_retrain}
    \resizebox{\columnwidth}{!}{
    \begin{tabular}{c|cccccccccccc  c}
    \toprule \#Finetuning & A$\rightarrow$C  & A$\rightarrow$P  & A$\rightarrow$R & C$\rightarrow$A & C$\rightarrow$P & C$\rightarrow$R & P$\rightarrow$A & P$\rightarrow$C & P$\rightarrow$R & R$\rightarrow$A & R$\rightarrow$C & R$\rightarrow$P & \textbf{Avg} \\
    \midrule
    0 & 85.0 & 96.2	& 93.9 & 89.4 & 96.8 & 94.1 & 87.5 & 85.4 & \textbf{94.5} & 90.2	& 85.3 & 97.1 & 91.3 \\
    1 (default) & 85.4 & 96.6 & 94.1 & 90.3 & 97.1 & 94.4 & 87.9 & 85.7 & \textbf{94.5} &90.1 &	85.5&	97.3 & 91.6\\
    2 & \textbf{85.6}	&96.8	&94.3	&90.4	&97.1&	94.7&	88.2&	\textbf{85.8}	&94.4&	89.9	&85.5&	\textbf{97.5}&	91.7\\
    3 & 85.5&	96.9&	94.5&	90.5&	97.2&	94.9&	\textbf{88.4}&	85.7&	\textbf{94.5}&	90.0&	85.6	&\textbf{97.5}&	\textbf{91.8}\\
    4 & 85.5	&\textbf{97.0}	&94.6	&90.2	&97.3&	\textbf{95.0}&	88.2&	85.7	&94.4&	90.3	&\textbf{85.8}	&\textbf{97.5}&	\textbf{91.8}\\
    5 & 85.4 	&\textbf{97.0} &	\textbf{94.8} &	\textbf{90.6} &	\textbf{97.4} &	94.9 &	88.3 &	85.6 	&\textbf{94.5}& 	\textbf{90.4}& 	\textbf{85.8} 	&\textbf{97.5} &	\textbf{91.8} \\
    \bottomrule
\end{tabular}}
\vskip -0.1 in
\end{table*}

In VLLaVO, we only finetune with pseudo-labels for only one time. Obviously, we can do that for multiple times. To identify the effect of the times of finetuning with pseudo-labels on the performance, Table \ref{tbl:uda_officehome_retrain} 
shows the testing accuracies with different times of finetuning with pseudo-labels for UDA tasks on the \textit{OfficeHome} dataset. According to the results, we can see that when the time increases from 0 to 3, the performance becomes better in terms of the average performance, and after that, the average performance becomes stable. Those results demonstrate the effectiveness of finetuning with pseudo-labels. However, with increasing times of finetuning with pseudo-labels, the time cost increases. So by default, we conduct only one finetuning with pseudo-labels.

\subsection{Effect of Instruction Template}
To verify the effectiveness of the designed question instruction template, we compare VLLaVO with its variants: VLLaVO (w/ domain information) and VLLaVO (w/ simple template). For VLLaVO (w/ domain information), we further consider the domain information in DG. Specifically, we add the domain name of each sample, denoted as \{\texttt{Domain Name}\}, to the designed question instruction. 
With the domain name, the template $\Tilde{\bT}(\vx)$ is shown in Example~\ref{ex:template2}.
For VLLaVO (w/ simple template), we use the template $\hat {\bT}(\vx)$ shown in Example~\ref{ex:template3}, which only employs the $\{\mathbb D (\vx)\}$ to formulate the question instruction.

\begin{exmp}{Template $\Tilde{\bT}(\vx)$.}{template2}
\label{exmp_template2}
Give the information about a \{\texttt{Domain Name}\} image:
\{$\mathbb{D}(\vx)$\}. \\
According to the information, choose the most similar category from the given options: 
\{\texttt{Category Set}\}. \\ 
\#\#\# Answer:
\end{exmp}

\begin{exmp}{Template $\hat{\bT}(\vx)$.}{template3}
\label{exmp_template3}
\{$\mathbb{D}(\vx)$\}.\\
\#\#\# Answer:
\end{exmp}

Table \ref{tbl:ablation_template} shows the testing accuracies of DG tasks on the \textit{OfficeHome} dataset by using the setting in Section \ref{sec:DG}.
As can be seen, similar cross-domain performance is achieved with or without the domain information. Therefore, for simplicity, we do not consider the domain information in VLLaVO.
Moreover, VLLaVO outperforms VLLaVO (w/ simple template), showing the rationality of our designed template. 

\begin{table}[!th]\small
\centering
\caption{Effects of instruction templates for DG tasks on the \textit{OfficeHome} dataset. The best is in \textbf{bold}.}
\label{tbl:ablation_template}
\begin{tabular}{cccccc}
    \toprule
Method & A & C & P & R & \textbf{Avg} \\
    \midrule
    VLLaVO  & 90.0 & \textbf{86.7} & \textbf{97.4} & 95.1 & 92.3 \\
    VLLaVO (w/ domain information) & \textbf{90.4} & 86.5 & 97.3 & \textbf{95.2} & \textbf{92.4} \\
    VLLaVO (w/ simple template) & 89.4 & 85.8 & 97.0 & 94.9 & 91.8 \\
    \bottomrule
\end{tabular}
\vskip -0.2in
\end{table}

\section{Released Text Datasets}
\label{sec:text_dataset}
\begin{table}[!th]\small
\vskip -.3in
\centering
\caption{Statistics of the released text datasets.} \label{table:ds_text} 
\vskip -.15in
\begin{tabular}{c c c c}
    \toprule
Dataset     & \#samples & \#classes & \#domains \\
    \midrule
    \textit{PACS}  & $9,991$ & $7$  & $4$ \\
    \textit{OfficeHome}  & $15,500$ & $65$ & $4$ \\
    \textit{DomainNet}  & $586,575$ & $345$  & $6$ \\
    \bottomrule
\end{tabular}
\vskip -0.3in
\end{table}

We release three datasets\footnote{They are available at \url{https://www.dropbox.com/scl/fo/brzjl4xzpfzf0r57tkkwr/h?rlkey=zxeabnjpkgbbpwugskpcjzrqa&dl=0}.} constructed from cross-domain image datasets, including 
\textit{PACS}, \textit{OfficeHome}, and \textit{DomainNet}. Each dataset contains multiple domains as in the corresponding image dataset. Each sample in a domain of one dataset contains the textual description, which is constructed in Section \ref{sec:extract}, for the corresponding image and the label is just the label of the corresponding image. Hence, each dataset released is for multi-domain text classification tasks. These datasets collectively contain 612,066 samples in total. 
Table \ref{table:ds_text} summarizes the statistics of those datasets, including the number of samples, classes, and domains. 

\section{Algorithms}
\label{sec:algorithm}
\begin{algorithm}[h]
\caption{Training pipeline of VLLaVO}
\label{alg:train_pipeline}
\renewcommand{\algorithmicrequire}{\textbf{Input:}} 
\renewcommand{\algorithmicensure}{\textbf{Output:}}
\begin{algorithmic}[1]
\REQUIRE Source domain $\mathcal{D}$, trainable LLM parameters $\vtheta$, description template $\mathbb{D}$, instruction template $\mathbb{T}$, learning rate $\eta$.
\STATE Initialize $\vtheta$;
\IF{Apply to UDA}
    \STATE $\hat \vtheta = \vtheta$;
\ENDIF
\REPEAT
    \STATE Sample an image $(\vx,y)$ from $\mathcal{D}$;
    \STATE Extract $\mathbb{D}(\vx)$ using VLMs via Section~\ref{sec:extract};
    \STATE Generate $\bT(\vx)$ using $\mathbb{D}(\vx)$;
    \STATE $\mathcal{L}_{dg}(\vtheta) = - \log p_{\vtheta} \big(y|\bT(\vx)\big)$;
    \STATE $\vtheta = \vtheta - \eta \nabla_\vtheta \mathcal{L}_{dg}(\vtheta)$;
\UNTIL{converged.}
\IF{Apply to UDA}
    \REPEAT
        \STATE Sample an image $\vx$ from $\mathcal{D} \cup \mathcal{T}$;
        \STATE Extract $\mathbb{D}(\vx)$ using VLMs via Section~\ref{sec:extract};
        \STATE Generate $\bT(\vx)$ using $\mathbb{D}(\vx)$;
        \IF{$\vx \in \mathcal{T}$}
            \STATE $\hat y =  LLM(\bT(\vx); \vtheta)$;
            \STATE $\mathcal{L}(\vtheta) = - \log p_{\hat \vtheta} \big(\hat y|\bT(\vx)\big)$;
            \STATE $\hat \vtheta = \hat \vtheta - \eta \nabla_{\hat \vtheta} \mathcal{L}(\vtheta)$;
        \ELSE
            \STATE $\mathcal{L}(\vtheta) = - \log p_{\hat \vtheta} \big(y|\bT(\vx)\big)$;
            \STATE $\hat \vtheta = \hat \vtheta - \eta \nabla_{\hat \vtheta} \mathcal{L}(\vtheta)$;
        \ENDIF
    \UNTIL{converged.}
    \STATE $\vtheta = \hat \vtheta$;
\ENDIF
\RETURN $\vtheta$
\end{algorithmic}
\end{algorithm}
The training pipeline of VLLaVO is summarized in Algorithm \ref{alg:train_pipeline}. 
To begin with, {VLLaVO} generates textual descriptions of images using pre-trained VLMs. After obtaining textual descriptions, we finetune an LLM with the designed question instruction template that combines the textual descriptions and their category labels. 
Moreover, when employing VLLaVO for UDA, the LLM is finetuned in the source domain to generate the pseudo-labels for target domain samples. The LLM is then finetuned with both target domain samples with pseudo-labels and source domain samples with ground truth labels.

\section{Limitation}
Despite showing promising results in various cross-domain tasks, the VLLaVO has certain limitations. 
Firstly, the quality of the extracted textual descriptions relies on the capabilities of the VLMs and there is still room for further improvement.
Examples of these textual descriptions can be found in Section~\ref{appendix:description_examples}. While most descriptions align well with the images, there are a few instances where the descriptions do not accurately match the images.
Secondly, our current work focuses solely on the visual classification task.
However, the domain shifts may also exist in the other visual tasks (\textit{e.g.}, segmentation, depth estimation, and surface normal prediction). 
Future research should explore methods to leverage the capabilities of LLMs in handling these types of visual cross-domain tasks.

\section{Case Study}
\subsection{More Examples of Textual Description}
\label{appendix:description_examples}
Example \ref{ex:exmp3}, \ref{ex:exmp4}, and \ref{ex:exmp5} show textual descriptions for images of different categories on the \textit{PACS} dataset, while Example \ref{ex:exmp6} and \ref{ex:exmp7} show textual descriptions for images of different categories on the \textit{OfficeHome} dataset. 
The Top-10 sensitive word (excluding stop words) is highlighted in red.
\begin{exmp}{Textual Description $\mathbb{D}(\vx)$ of Image $\vx$.}{exmp3}
    \footnotesize
    \vspace{-0.03in}
    \begin{minipage}[]{0.80\textwidth}
    {\small \textbf{Tags:}} \\
    -\textcolor{red}{giraffe} \\
    -\textcolor{red}{Giraffidae} \\
    -Watercolor paint \\
    -\textcolor{red}{Camelid} \\
    -Animal product 
    \end{minipage}%
    \begin{minipage}[]{0.20\textwidth}
    \raggedleft
    \includegraphics[height=0.9in]{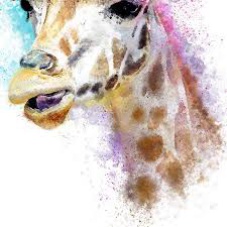}
    \vspace{-0.23in}
    \begin{center}
        (label:~giraffe)
    \end{center}
    \end{minipage}
    {\small \textbf{Attributes}}: \\
    -\textcolor{red}{giraffe} which has dark spots on the coat \\
    -\textcolor{red}{giraffe} which has long neck \\
    -\textcolor{red}{giraffe} which has cloven hooves \\
    -animal which has muzzle \\
    -artwork which is a creative work of art, such as a painting, sculpture, drawing, or photograph \\
    
    {\small \textbf{Captions}}: \\
    -a \textcolor{red}{giraffe} with pink colored hair walking \\
    -a closeup view of a \textcolor{red}{giraffe} with a very very painted look on his face \\
    -a painting of a \textcolor{red}{giraffe} with its tongue hanging out \\
    -the \textcolor{red}{giraffe} is wearing a crown and with his tongue out \\
    -a watercolor painting of a \textcolor{red}{giraffe} with a tongue out \\
    -a watercolor portrait of an adorable \textcolor{red}{giraffes} face \\
    -a \textcolor{red}{giraffe} staring at the camera with a paint splatter effect \\
    -a painting of a \textcolor{red}{giraffe} on a wall \\
    -an artistic picture of a \textcolor{red}{giraffe} with a colorful head piece \\
    -an image of a \textcolor{red}{giraffe} that is on a canvas 
\end{exmp}

\begin{exmp}{Textual Description $\mathbb{D}(\vx)$ of Image $\vx$.}{exmp4}
    \footnotesize
    \vspace{-0.03in}
    \begin{minipage}[]{0.80\textwidth}
    {\small \textbf{Tags:}} \\
    -Sled \textcolor{red}{dog} \\
    -Alaskan \textcolor{red}{malamute} \\
    -Siberian \textcolor{red}{husky} \\
    -Seppala siberian \textcolor{red}{sleddog} \\
    -Sakhalin \textcolor{red}{husky} 
    \end{minipage}%
    \begin{minipage}[]{0.20\textwidth}
    \raggedleft
    \includegraphics[height=0.9in]{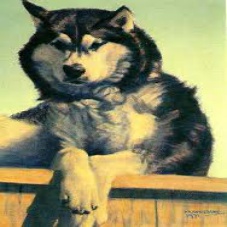}
    \vspace{-0.23in}
    \begin{center}
        (label:~dog)
    \end{center}
    \end{minipage}
    {\small \textbf{Attributes}}: \\
    -alaskan \textcolor{red}{malamute} which has black, grey, white, or sable coat with a mask-like pattern on the \textcolor{red}{face} \\
    -siberian \textcolor{red}{husky} which has dense undercoat with a longer, thicker outer coat \\
    -alaskan \textcolor{red}{malamute} which has broad head \\
    -alaskan \textcolor{red}{malamute} which has thick double coat of fur \\
    -siberian \textcolor{red}{husky} which has thick double coat of fur \\
    
    {\small \textbf{Captions}}: \\
    -a painting of a \textcolor{red}{husky} \textcolor{red}{dog} with a \textcolor{red}{dog} bone on its paw \\
    -a painting of a \textcolor{red}{dog} on a wooden deck \\
    -an alaska wolf \textcolor{red}{dog} is \textcolor{red}{sitting} on a fence \\
    -a black and white \textcolor{red}{husky} \textcolor{red}{dog} \textcolor{red}{sitting} in a bath tub \\
    -a painting of a wolf laying on a ledge \\
    -a \textcolor{red}{dog} \textcolor{red}{sitting} on top of a box next to a wall \\
    -a cover of a book with a painting of a \textcolor{red}{husky} \textcolor{red}{dog} \\
    -a \textcolor{red}{dog} \textcolor{red}{sitting} on a ledge looking into the distance \\
    -a \textcolor{red}{dog} on top of a wooden wall with green sky in background \\
    -a picture of a huge \textcolor{red}{dog} \textcolor{red}{sitting} down on the ledge \\
\end{exmp}


\vspace{-0.25in}
\begin{exmp}{Textual Description $\mathbb{D}(\vx)$ of Image $\vx$.}{exmp5}
    \footnotesize
    \vspace{-0.03in}
    \begin{minipage}[]{0.80\textwidth}
    {\small \textbf{Tags:}} \\
    -\textcolor{red}{Mustang} \textcolor{red}{horse} \\
    -\textcolor{red}{horse} \\
    -\textcolor{red}{stallion} \\
    -\textcolor{red}{forelock} \\
    -foal 
    \end{minipage}%
    \begin{minipage}[]{0.20\textwidth}
    \raggedleft
    \includegraphics[height=0.9in]{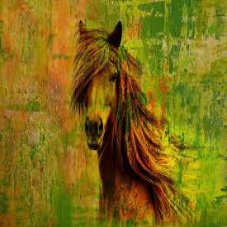}
    \vspace{-0.23in}
    \begin{center}
        (label:~horse)
    \end{center}
    \end{minipage}
    {\small \textbf{Attributes}}: \\
    -\textcolor{red}{horse} which has mane of hair on neck \\
    -\textcolor{red}{stallion} which has powerful build \\
    -animal which has whiskers or mane
    -artwork which is a creative work of art, such as a painting, sculpture, drawing, or photograph \\
    -\textcolor{red}{horse} which has furry body \\
    
    {\small \textbf{Captions}}: \\
    -a brown \textcolor{red}{horse} with its long mane running in the field \\
    -a painting of a big brown \textcolor{red}{horse} with long mane \\
    -a picture of a \textcolor{red}{horse} standing on a field \\
    -an orange - colored \textcolor{red}{horse} walking across a green, yellow and yellow - green grassy area \\
    -a painting of two horses in a field
    -the \textcolor{red}{horse's} head is shown with long hair hanging \\
    -horses standing near each other in some grass \\
    -a painting of a \textcolor{red}{horse} that is standing up by itself \\
    -painting of a \textcolor{red}{horse} with a shaggy hair blowing in the wind \\
    -a painting of a \textcolor{red}{horse} in a green field 
\end{exmp}

\vspace{-0.5in}
\begin{exmp}{Textual Description $\mathbb{D}(\vx)$ of Image $\vx$.}{exmp6}
    \footnotesize
    \vspace{-0.03in}
    \begin{minipage}[]{0.72\textwidth}
    {\small \textbf{Tags:}} \\
    -saddlebag \\
    -Road \textcolor{red}{bicycle} \\
    -\textcolor{red}{bicycle} \\
    -Conference \textcolor{red}{Bike} \\
    -Tandem \textcolor{red}{bicycle} \\
    \end{minipage}%
    \begin{minipage}[]{0.28\textwidth}
    \raggedleft
    \includegraphics[height=0.9in]{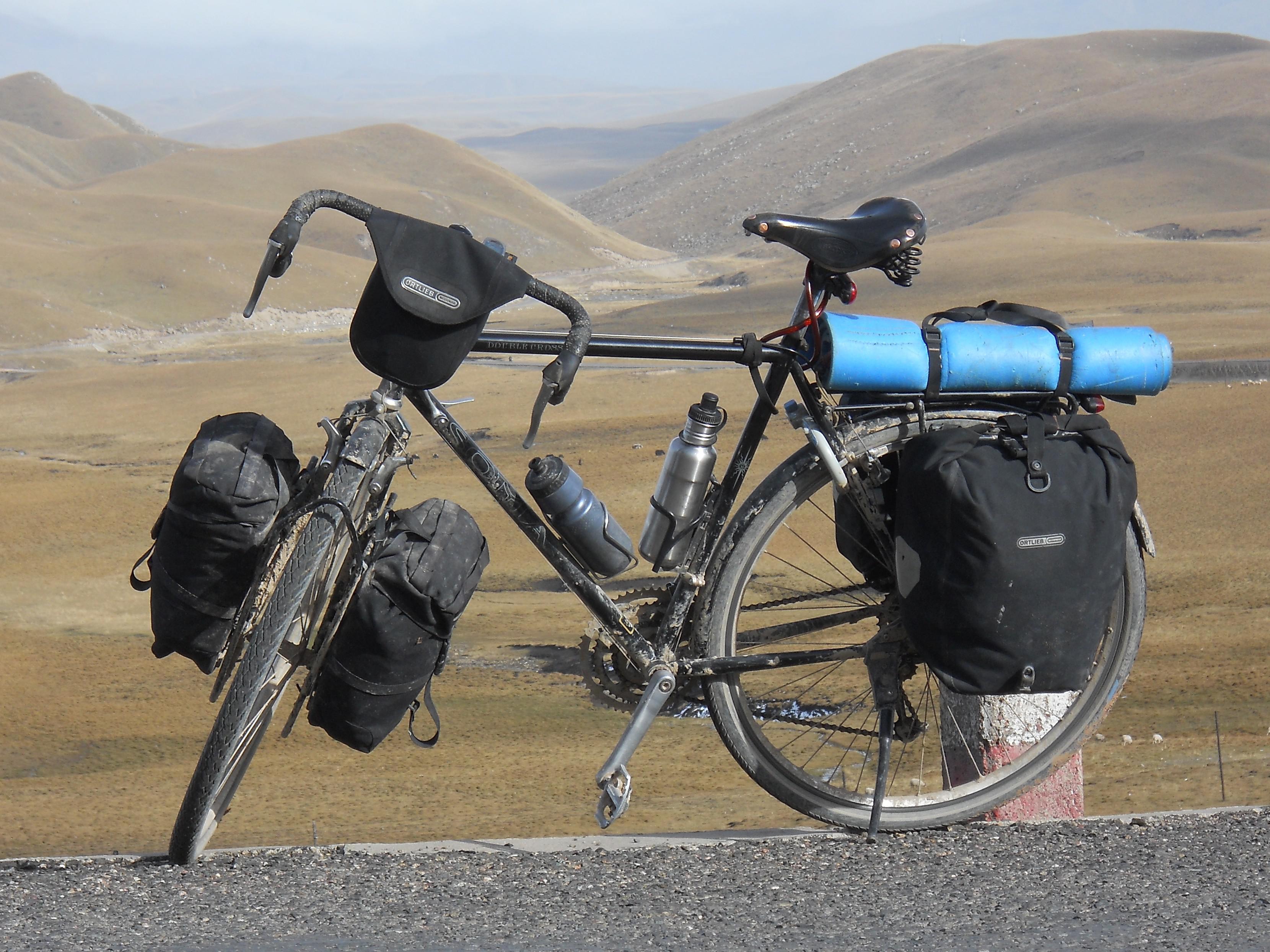}
    \vspace{-0.15in}
    \begin{center}
        (label:~bike)
    \end{center}
    \end{minipage}
    {\small \textbf{Attributes}}: \\
    -cyclist which is a saddle bag \\
    -\textcolor{red}{bike} which is a kickstand \\
    -mountain \textcolor{red}{bike} which has kickstand \\
    -mountain \textcolor{red}{bike} which has straight handlebars \\
    -\textcolor{red}{bike} which has handlebars \\
    
    {\small \textbf{Captions}}: \\
    -two backpacks and \textcolor{red}{bicycles} parked next to each other \\
    -a \textcolor{red}{bicycle} with a bundle of luggage and \textcolor{red}{bags} parked at the side of a road \\
    -a \textcolor{red}{bicycle} that is sitting in the grass \\
    -a couple of bikes with \textcolor{red}{bags} strapped on a \textcolor{red}{bicycle} rack \\
    -a \textcolor{red}{bike} has two wheels and has some \textcolor{red}{bags} \\
    -a \textcolor{red}{bicycle} with a blue sack on the back and saddle attached \\
    -a \textcolor{red}{bike} that is sitting next to the mountains \\
    -a \textcolor{red}{bike}, loaded with a blue backpack, parked on a road \\
    -a \textcolor{red}{bicycle} with two \textcolor{red}{bags} that include backpacks \\
    -\textcolor{red}{bicycle} with three \textcolor{red}{bags} on the back rack \\
\end{exmp}

\begin{exmp}{Textual Description $\mathbb{D}(\vx)$ of Image $\vx$.}{exmp7}
    \footnotesize
    \vspace{-0.03in}
    \begin{minipage}[]{0.82\textwidth}
    {\small \textbf{Tags:}} \\
    -Tin \textcolor{red}{can} \\
    -\textcolor{red}{pail} \\
    -Oyster \textcolor{red}{pail} \\
    -Watering \textcolor{red}{can} \\
    -cannister \\
    \end{minipage}%
    \begin{minipage}[]{0.18\textwidth}
    \raggedleft
    \includegraphics[height=0.9in]{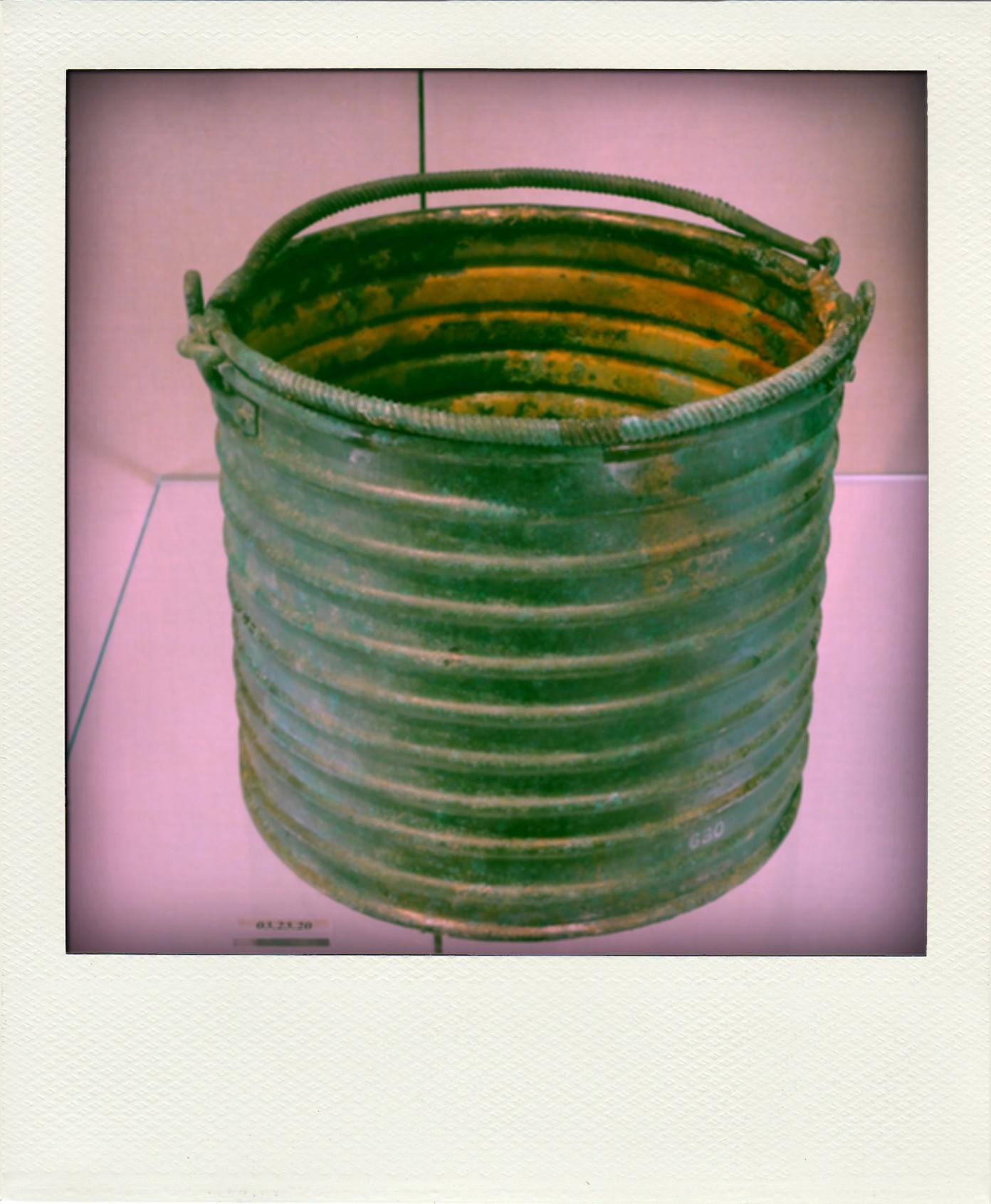}
    \vspace{-0.15in}
    \begin{center}
        (label:~bucket)
    \end{center}
    \end{minipage}
    {\small \textbf{Attributes}}: \\
    -winebucket which has cylindrical or oval shape \\
    -\textcolor{red}{bucket} which has cylindrical container \\
    -\textcolor{red}{pail} which is a cylindrical or \textcolor{red}{bucket \\}-shaped container \\
    -\textcolor{red}{pail} which has usually has a lid \\
    -\textcolor{red}{pail} which has usually has a handle\\
    
    {\small \textbf{Captions}}: \\
    -a green and rusty metal \textcolor{red}{bucket} \textcolor{red}{sitting} on a \textcolor{red}{table} \\
    -a close up of a metal basket on a \textcolor{red}{table} \\
    -a picture of a \textcolor{red}{bucket} filled with dirt \\
    -an object with a holed handle on a small cup \\
    -an empty green pot in a pink room \\
    -a basket \textcolor{red}{sitting} in a room near a wall \\
    -a large pot \textcolor{red}{sitting} on a counter \\
    -a \textcolor{red}{bucket} that is green \textcolor{red}{sitting} on a \textcolor{red}{table} \\
    -a green pot on a \textcolor{red}{table} on a pink background \\
    -a rusted tin \textcolor{red}{bucket} \textcolor{red}{sitting} on a pink \textcolor{red}{table} \\
\end{exmp}

\subsection{Examples of Undesirable Answers}
\label{appendix:unpredictable_content}

To explore the reason behind the limited performance of ZS-LLM, in Table \ref{tbl:undesirable} we present a list of undesirable answers with the `person' category on the \textit{PACS} dataset. 
Specifically, we feed $\bT(x)$ into the pretrained LLaMA and generate the corresponding output with beam search.  
As can be seen, undesirable answers can be categorized into two groups: (1) The answer is correlated to the label but represented in an undesirable way. 
It includes extraneous punctuation marks like ``~'~'' and ``~[~'' as well as phrases such as ``The most similar category to the given information'' and reasons for classification. 
Those undesirable answers may be due to that the training objective in the next word prediction task during the pretraining process (\textit{e.g.}, ``The most similar'') is still misaligned with the expected zero-shot learning objective (\textit{e.g.}, ``person''). 
(2) Undesirable answers do not belong to the provided options. For instance, the answers in the last three rows mention ``woman'', ``man'', and ``movie''. 
Those answers show that the pretrained language model is limited in its ability to follow instructions.

\newcolumntype{P}[1]{>{\raggedright\arraybackslash}p{#1}}
\begin{table}[!t]\small
\centering
\caption{Example of undesirable answers on the \textit{PACS} dataset. }
\label{tbl:undesirable}
\vskip -.15in
\linespread{2}
\begin{tabular}{P{1.2cm}P{5cm}}
    \toprule
    \textbf{Label} & \textbf{Answer} \\
    \midrule
    person & `person' \\  \midrule
    person & [`person'] \\  \midrule
    person & The most similar category to the given information is `person'. \\  \midrule
    person & The most similar category from the given options is `person'.\\   \midrule
    person & The most similar category based on the information provided is [`person']. \\ \midrule
    person & The most similar category to the given information is `person'. The image is of an astronaut in outer space, which is a subject or theme related to `person'. \\  \midrule
    person & The most similar category to the given information is ``person''. The image is a painting of a person, specifically a man jumping over a crowd, standing over a bunch of people, or running with a group of others.\\ \midrule
    person & Given the information provided, the most similar category to the images is ``person''. This is because all of the images depict human figures, either as the main subject or as part of a larger scene. Therefore, the category ``person''. \\ \midrule
    person & The most similar category to the given information is `woman'. \\ \midrule
    person & The most similar category to the given information is `man'. \\ \midrule
    person & The most similar category to the given information is [`movie']. \\ \bottomrule
\end{tabular}
\end{table}

\end{document}